\documentclass[preprint,authoryear,12pt]{elsarticle}

\usepackage{amssymb}
\usepackage{graphicx}
\usepackage[fleqn]{amsmath}
\usepackage{subfigure}
\usepackage{multirow}

\renewcommand{\Vec}[1]{ \mbox{\boldmath $#1$} }
\renewcommand{\matrix}[2]{ \left[ \begin{array}{#1} #2 \end{array} \right] }
\newcommand{\tr}[1]{ {\rm tr}\left[ #1 \right] }
\newcommand{\pnorm}[2]{ \left\| #1 \right\|_{#2}}
\newcommand{\snorm}[1]{ \left\| #1 \right\|_{\rm S}}
\newcommand{\diag}[1]{ {\rm diag}\left( #1 \right) }
\newcommand{\bdiag}[1]{ {\rm block-diag}\left( #1 \right) }

\newcommand{\argmin}[1]{ \underset{#1}{\rm argmin} }
\newcommand{\argmax}[1]{ \underset{#1}{\rm argmax} }
\newcommand{\proj}[2]{ {\rm proj} \left( #1, #2 \right) }

\newcommand{\R}{ \mathbb{R} }
\newcommand{\ND}{ \mathcal{N} }
\newcommand{\PD}{ \mathcal{P} }
\newcommand{\Order}{ \mathcal{O} }
\newcommand{\Lag}{ \mathcal{L} }
\newcommand{\Lm}{ \Lambda }
\newcommand{\Th}{ \Theta }
\newcommand{\Om}{ \Omega }
\newcommand{\C}{ \mathcal{C} }
\newcommand{\pC}{ \partial \mathcal{C} }
\newcommand{\KL}{D_{\text{KL}}}

\newcommand{\gm}{ \gamma }

\newcommand{\sg}{ \sigma }
\newcommand{\tsg}{ {\tilde{\sigma}} }

\newcommand{\indep}{\mathop{\perp\!\!\!\perp}}
\newcommand{\sgn}[1]{ {\rm sgn} \left( #1 \right) }
\newcommand{\st}{ \; {\rm s.t.} \;\; }

\journal{Neural Networks}

\begin{document}

\begin{frontmatter}



\title{Learning a Common Substructure of Multiple Graphical Gaussian Models}


\author[add1]{Satoshi Hara}
\author[add1]{Takashi Washio}
\address[add1]{Institute of Scientific and Industrial Research (ISIR), Osaka University, Osaka, 5670047, Japan}

\begin{abstract}

Properties of data are frequently seen to vary depending on the sampled situations, 
which usually changes along a time evolution or owing to environmental effects.
One way to analyze such data is to find invariances, or representative features kept constant over changes.
The aim of this paper is to identify one such feature, namely interactions or dependencies among variables 
that are common across multiple datasets collected under different conditions.
To that end, we propose a common substructure learning (CSSL) framework based on a graphical Gaussian model.
We further present a simple learning algorithm based on the Dual Augmented Lagrangian and the Alternating Direction Method of Multipliers.
We confirm the performance of CSSL over other existing techniques in finding unchanging dependency structures in multiple datasets 
through numerical simulations on synthetic data and through a real world application to anomaly detection in automobile sensors.
\end{abstract}


\end{frontmatter}

\newtheorem{definition}{Definition}
\newtheorem{theorem}{Theorem}
\newtheorem{lemma}{Lemma}
\newtheorem{proposition}{Propostion}
\newtheorem{corollary}{Corollary}

% \linenumbers

%\section{}
%\label{}

%==============================================================================================
\section{Introduction}

In several real world data, such as that from the stock market~\citep{baillie1989common}, 
gene regulatory networks~\citep{ahmed2009recovering,zhang2009differential}, 
biomedical measurements~\citep{varoquaux2010brain}, or sensors in engineering systems~\citep{ide2009proximity}, 
there are dynamical properties over time evolutions or due to changes in the surrounding environments.
Such effects cause data to have different behaviors in each dataset collected under different conditions.
One way to analyze such data is to explicitly include the change into the model~\citep{hamilton1994time,durbin2001time}, 
which usually requires detailed domain knowledge that is rarely available in most cases.
Another way is to impose general and mild assumptions on the data.
This kind of approach is especially common in the multi-task learning literatures~\citep{caruana1997multitask,turlach2005simultaneous}, 
where the relationships among datasets are treated as a clue for combining multiple tasks into a single problem. 
The scope of the present paper is in the latter context where the relationship among datasets is the objective we want to analyze.
For the purpose, we focus on invariance of the data against the underlying changes 
which provides partial yet important aspects of the data behaviors~\citep{von2009finding,hara2012separation}.
We provide a technique for finding one of such invariance, 
specifically constant interactions or dependencies among variables across several different conditions.
An illustrative example is an engineering system where system errors are observed as dependency anomalies in sensor values~\citep{ide2009proximity}, 
which are usually caused by a fault in a subsystem.
The invariance, which in this example is the remaining healthy subsystems, is captured by a steady dependency over
the multiple datasets sampled before and after the error onset.
Hence, we can use such information as a clue for finding erroneous subsystems.

Graphical modeling is a popular approach for analyzing dependencies in multivariate data~\citep{lauritzen1996graphical}.
We adopt one of the most fundamental models, a graphical Gaussian model (GGM), as the basis of our framework.
A GGM is a basic model representing {\it linear} dependencies among {\it continuous} random variables, 
and has been widely studied owing to the simple nature, that is, 
the dependency structure is represented by the zero patterns in an inverse covariance matrix.
Identification of such zero patterns from data was first studied by~\cite{dempster1972covariance} as a {\it Covariance Selection}
where the task is formulated as the combinatorial problem of optimizing the location of zeros in a matrix.
Since classical algorithms for this do not scale to high dimensional data, 
the scope of studies has shifted to a relaxed setting~\citep{meinshausen2006high,yuan2007model,banerjee2008model}, 
where Covariance Selection is formulated as a convex optimization problem using a $\ell_1$-regularization that induces zeros in the resulting matrix.
Because of the effectiveness of the relaxed formulation, several related optimization techniques have also been
studied~\citep{friedman2008sparse,duchi2008projected,li2010inexact,scheinberg2010learning,yuan2009alternating,scheinberg2010sparse,hsiehsparse}.

In our context, the objective is not to estimate the structure of a GGM from a single dataset, 
but to decompose the resulting GGMs from several datasets into common and individual substructures, 
with the former representing the invariance we aim to detect.
There are some prior studies on learning a set of GGMs from multiple datasets.
\cite{varoquaux2010brain} and \cite{honorio2010multi} imported the idea of Group-Lasso~\citep{yuan2006model,bach2008consistency}
and Multitask-Lasso~\citep{turlach2005simultaneous,liu2009blockwise}, and extended the framework of a single GGM setting.
In both cases, the problem is formulated under the assumption that all matrices share the same zero patterns.
\cite{guo2011joint} considered a method to avoid this additional assumption, although the problem then loses convexity.
Though these approaches achieved some success in improving the estimation accuracy of graphical models, 
this does not necessarily mean that they are suitable for finding commonness across datasets as we will see in the simulation.
In the context of common substructure detection, \cite{zhanglearning} proposed using a Fused-Lasso~\citep{tibshirani2005sparsity} 
type of technique to find an invariant pattern between two datasets.
As a general framework for $N$ datasets situations, \cite{chiquet2011inferring} considered imposing sign coherence on the resulting structures, 
while \cite{hara2011common} extended the framework of \cite{zhanglearning} to the general situation of 
$N$ datasets~\footnote{This paper is an extension of \citet{hara2011common} with more general settings, an efficient optimization algorithm, 
and exhaustive simulations on synthetic and real world datasets.}.
In the opposite context where the target is dynamics rather than invariance, 
\cite{zhou2010time} proposed using weighted statistics to trace the evolution of a GGM.
We note there are also several related studies in the binary Markov random field literatures~\citep{guo2007recovering,ahmed2009recovering}.
They also use $\ell_1$-regularization~\citep{wainwright2007high} and Fused-Lasso type techniques~\citep{ahmed2009recovering}
for recovering temporal dependency structures, which are technically quite close to the ones of GGM.

The contribution of this paper is two folds.
First, we introduce the novel {\it Common Substructure Learning} (CSSL) framework that is applicable for a general case of $N$ datasets.
Second, a sophisticated algorithm based on the Dual Augmented Lagrangian (DAL)~\citep{tomioka2011super} 
and the Alternating Direction Method of Multipliers (ADMM)~\citep{gabay1976dual,boyd2011distributed} is proposed.
In the proposed algorithm, the inner problems for each iterative update are simple and can be solved efficiently which results in fast computation.
We confirm the validity of the CSSL approach through simulations on synthetic datasets and on an anomaly detection task in real-world data.

The remainder of the paper is organized as follows.
In Section~\ref{sec:ggm}, we briefly review properties of GGMs and existing learning techniques.
In Section~\ref{sec:cssl}, we present the proposed framework and its theoretical properties.
The optimization algorithm based on DAL-ADMM is introduced in Section~\ref{sec:alg}.
The validity of the proposed method is presented through synthetic experiments in Section~\ref{sec:simu}.
In Section~\ref{sec:anom}, we apply the proposed method to an anomaly detection task on {\it sensor error} data.
Finally, we conclude the paper in Section~\ref{sec:concl}.

%==============================================================================================
\section{Structure Learning of Graphical Gaussian Model}
\label{sec:ggm}

\begin{table}[t]
\centering
%\caption{Mathematical Notations}
\caption{Mathematical Notation}
\label{tab:notation}
\begin{tabular}{cl}
	\hline
	\hspace{-4pt} Notation \hspace{-4pt} & Description \\
	\hline
	\hspace{-6pt} $\pnorm{\Vec{x}}{p}$ \hspace{-4pt}
		& $\ell_p$-norm of a vector $\Vec{x} \in \R^d$, $\pnorm{\Vec{x}}{p} = \left( \sum_{i=1}^d |x_i|^p \right)^\frac{1}{p}$ \\
		& for $p \in [1, \infty)$ and $\pnorm{\Vec{x}}{\infty} = \max_{1 \leq i \leq d} |x_i|$ \\
	\hspace{-6pt} $\pnorm{A}{p}$ \hspace{-4pt}
		& vectorized $\ell_p$-norm of a matrix $A \in \R^{d \times d}$, \\
		& $\pnorm{A}{p} = \pnorm{(A_{11}, A_{12}, \ldots, A_{dd})^\top}{p}$ \\
	\hspace{-6pt} $\snorm{A}$ \hspace{-4pt}
		& spectral norm of a matrix $A \in \R^{d \times d}$, \\
		& $\snorm{A} = \max_{1 \leq i \leq d} \sigma_i(A)$ where $\sigma_i(A)$ is \\
		& an $i$th singular value of $A$ \\
	\hspace{-6pt} $\pnorm{B}{1, p}$ \hspace{-4pt}
		& $\ell_{1, p}$-norm of matrices $B = \{ B_i ; B_i \in \R^{d \times d}\}_{i=1}^N$, \\
		& $\pnorm{B}{1, p} = \sum_{j, j' = 1}^d \pnorm{(B_{1, jj'}, B_{2, jj'}, \ldots, B_{N, jj'})^\top}{p}$ \\
	\hspace{-6pt} $A \succ 0$ \hspace{-4pt}
		& a matrix $A$ is symmetric and positive definite \\
	\hspace{-6pt} $\sgn{a}$ \hspace{-4pt}
		& sign function on a scalar $a$, $\sgn{a} = 1$ for $a > 1$, \\
		& $\sgn{a} = -1$ for $a < 0$ and $\sgn{a} = 0$ for $a = 0$ \\
	\hspace{-6pt} $\diag{\Vec{x}}$ \hspace{-6pt}
		& $d \times d$ matrix with $\Vec{x} \in \R^d$ on its diagonal \\
	\hline
\end{tabular}
\vspace{-2.5ex}
\end{table}

In this section, we review the GGM estimation problem~\citep{meinshausen2006high,yuan2007model,banerjee2008model}
and some prior extensions to multiple datasets \citep{varoquaux2010brain,honorio2010multi,zhanglearning}.

We also summarize mathematical notations used throughout the paper in \tablename~\ref{tab:notation}.

%==============================================================================================
\subsection{Graphical Gaussian Model}
\label{subsec:ggm}

In multivariate analysis, covariance and correlation are commonly used as indicators for a relationship between two random variables.
However, in general, a covariance between two random variables $x_j$ and $x_{j'}$ is affected by other variables.
Therefore, we need to remove such effects to estimate an essential dependency structure, 
which is available by searching for conditional dependency among random variables.
In a general graphical model, we express these dependencies using a graph with vertices corresponding to each random variable and 
edges spanning random variables that are conditionally dependent.

Here, we assume that a $d$-dimensional random variable $\Vec{x} = (x_1, x_2, \ldots, x_d)^\top$ follows a zero mean Gaussian distribution, 
that is, $\Vec{x} \sim \ND(\Vec{0}_d, \Lm^{-1})$ for some symmetric and strictly positive definite matrix $\Lm \in \R^{d \times d}$.
We refer to a graphical model of Gaussian variables as graphical Gaussian model (GGM)
Note that the zero mean assumption can be achieved without loss of generality by subtracting a sample mean from the dataset.
Here, a covariance matrix is parameterized as the inverse of a {\it precision matrix} $\Lm$ since 
this is a more primitive parameter representing essential dependency among variables.
A precision matrix relates to the conditional expectation as
\begin{align*}
	\Lm_{jj'} \propto - \mathbb{E} \left[ x_j x_{j'} | \text{other variables} \right] \, ,
\end{align*}
that is, the $(j, j')$th entry of $\Lm$ is proportional to the covariance between $x_j$ and $x_{j'}$ with the remaining $d-2$ variables fixed.
With this property, the conditional independence between Gaussian random variables is expressed as zero entries of $\Lm$:
\begin{align*}
	\Lm_{jj'} = 0 \; \Leftrightarrow \; x_j \indep x_{j'} \; | \; \text{other variables}
\end{align*}
where $\indep$ denotes statistical independence.
Because of this property, the edge patterns in a GGM correspond to the non-zero entries in a precision matrix $\Lm$.
In a GGM, two vertices have an edge between them if and only if the corresponding $(j, j')$th entry of $\Lm$ is non-zero.
In the case that only few pairs of variables are dependent, most off-diagonal elements in $\Lm$ are zeros 
and the corresponding graph expression is sparse, which allows us to visually inspect the underlying relations.

%==============================================================================================
\subsection{Sparse Estimation of GGM}
\label{subsec:glasso}

A naive way to estimate a precision matrix $\Lm$ is a maximum likelihood estimation formulated as
\begin{align}
	& \hat{\Lm} = \argmax{\Lm \in \PD} \; \ell(\Lm ; S) \, , \nonumber \\
	& \ell(\Lm ; S) = \log \det \Lm - \tr{S \Lm} \, .
	\label{eq:ml}
\end{align}
Here, $\ell(\Lm ; S)$ is a log-likelihood of a Gaussian distribution (up to a constant), $S$ is a sample covariance matrix
and $\PD$ is a set of symmetric positive definite matrices $\PD = \{ A \in \R^{d \times d} ; A \succ 0 \}$.
The positive definiteness constraint is imposed so that the resulting $\Lm$ is a valid precision matrix.
For a strictly positive definite matrix $S$, the solution to this problem is $\hat{\Lm} = S^{-1}$.
However, in a finite sample case, even when the true parameter is zero, that is, $\Lm_{jj'} = 0$, 
its maximum likelihood estimator $\hat{\Lm}_{jj'}$ is non-zero with probability one.
In this situation, the resulting graphical model is a complete graph, which states that every pairs of variables is conditionally dependent 
and the underlying intrinsic relationships are masked.

The major scope of GGM studies is how to avoid this unfavorable result from a maximum likelihood estimation and infer a sparse graph structure, 
which is referred to as {\it Covariance Selection} \citep{dempster1972covariance}.
In classical studies, some entries of a precision matrix $\Lm$ are fixed as zeros and the remaining non-zero entries are estimated, 
where the zero pattern is optimized in a combinatorial manner.
However, this combinatorial problem is not feasible for high-dimensional data.

In recent studies, the use of an $\ell_1$-regularization has been shown to be practical for Covariance Selection.
The first such study was conducted by \cite{meinshausen2006high}.
In their approach, the solution is obtained by solving the Lasso \citep{tibshirani1996regression}.
Here, let us denote $d$-dimensional data with $n$ data points using an $n \times d$ matrix 
$X = \matrix{cccc}{\Vec{x}_1 & \Vec{x}_2 & \ldots & \Vec{x}_n}^\top$, 
with $X_j$ as its $j$th column and $X_{\setminus j}$ as its remaining $d-1$ columns.
For each column, we solve the following Lasso:
\begin{align}
	\min_{\Vec{\theta}} \; \frac{1}{2} \pnorm{X_j - X_{\setminus j} \Vec{\theta}}{2}^2 + \rho \pnorm{\Vec{\theta}}{1} \, ,
	\label{eq:lasso}
\end{align}
where $\rho \geq 0$ is a regularization parameter. 
We then set zero patterns of $\Vec{\theta}$ to the $j$th column of $\Lm$.
\citet{meinshausen2006high} have also showed the asymptotic convergence of their estimator to the true graph structure under a proper condition.
This approach was later reformulated as an $\ell_1$-regularized maximum likelihood problem~\citep{yuan2007model,banerjee2008model}: 
\begin{align}
	\max_{\Lm \in \PD} \; \ell(\Lm ; S) - \rho \pnorm{\Lm}{1} \, .
	\label{eq:sics}
\end{align}
We refer to this problem as {\it Sparse Inverse Covariance Selection} (SICS) following \cite{scheinberg2010sparse}.
The resulting precision matrix of (\ref{eq:sics}) has some zero entries owing to the effect of an additional $\ell_1$-regularization term.
Several efficient optimization techniques are available for solving this problem.
Examples include GLasso \citep{friedman2008sparse}, PSM \citep{duchi2008projected}, IPM \citep{li2010inexact}, 
SINCO \citep{scheinberg2010learning}, ADMM \citep{yuan2009alternating,scheinberg2010sparse} and QUIC \citep{hsiehsparse}.

%==============================================================================================
\subsection{Learning a Set of GGMs with Same Topological Patterns}
\label{subsec:joint_sics}

The ordinary SICS problem~(\ref{eq:sics}) aims to learn one GGM from a single dataset.
The extension of this framework to multiple datasets has been studied by \cite{varoquaux2010brain} and \cite{honorio2010multi}.
The task is to estimate $N$ precision matrices $\Lm_1, \Lm_2, \ldots, \Lm_N$ from $N$ datasets
where the sample covariance matrices for each dataset are $S_1, S_2, \ldots, S_N$.
The objective of this multi-task extension is to improve the estimation accuracy of each GGM by incorporating the similarity among datasets.
In the framework of the above studies, GGMs from each dataset are assumed to have the same topological patterns, 
that is, the same edge connection structures while the edge weights might be different for each GGM.
They both introduced a $\ell_{1, p}$-norm of a set of $N$ precision matrices $\{ \Lm_i \}_{i=1}^N$
\begin{align*}
	\pnorm{\Lm}{1, p} = \sum_{j, j' = 1}^d \left( \sum_{i=1}^N \left| \Lm_{i, jj'} \right|^p \right)^\frac{1}{p} \, ,
\end{align*}
as a regularization term
analogous to the Group-Lasso~\citep{yuan2006model,bach2008consistency} and Multitask-Lasso~\citep{turlach2005simultaneous,liu2009blockwise}
with $p \in [1, \infty]$.
\citet{varoquaux2010brain} has considered the case $p = 2$ while \citet{honorio2010multi} used $p = \infty$.
These two choices are commonly adopted in many scenarios owing to the computational efficiency.
The entire estimation problem is defined as
\begin{align}
	\max_{ \left\{ \Lm_i ; \Lm_i \in \PD \right\}_{i=1}^N } \sum_{i=1}^N t_i \ell(\Lm_i ; S_i) - \rho \pnorm{\Lm}{1, p} \, ,
	\label{eq:msics}
\end{align}
with non-negative weights $t_1, t_2, \ldots, t_N$.
Without loss of generality, we can limit ourselves to the normalized case $\sum_{i=1}^N t_i = 1$ 
since the unnormalized version is just a scaled objective function for some constant.
The typical choice of parameters is $t_i = \frac{n_i}{\sum_{i=1}^N n_i}$ where $n_i$ is the number of data points in the $i$th dataset.
We refer to problem~(\ref{eq:msics}) as {\it Multitask Sparse Inverse Covariance Selection} (MSICS) in the remainder of the paper.

Note that the MSICS problem~(\ref{eq:msics}) involves the ordinary SICS~(\ref{eq:sics}) as a special case when $p=1$
where the $\ell_{1, 1}$-regularization term completely decouples into $N$ individual $\ell_1$-regularizations.
In the extended case for $p > 1$, the regularization term enforces the joint structure 
$\tilde{\Lm}_{jj'} = \left( \sum_{i=1}^N |\Lm_{i, jj'}|^p \right)^\frac{1}{p}$ to be sparse, 
with $\tilde{\Lm}_{jj'} = 0$ indicating that the corresponding $(j, j')$th entries are zeros across all $N$ precision matrices.

%==============================================================================================
\subsection{Learning Structural Changes between Two GGMs}
\label{subsec:fused_glasso}

Although taking advantage of situations with multiple datasets using the preceding techniques is useful
for improving the estimation performances of the resulting GGMs, 
it only imposes joint zero patterns and does not indicate anything about the commonness of the non-zero entries.
It is therefore not that helpful when comparing GGMs representing similar models 
where we expect that there may exist some common edges whose weights are close to each other.
\cite{zhanglearning} considered the two datasets case and constructed an algorithm
using a Fused-Lasso type regularization~\citep{tibshirani2005sparsity} to round these similar values to be exactly the same
allowing only significantly different edges between two GGMs to be extracted.
Their approach follows the ideas of \cite{meinshausen2006high} by connecting the update procedure (\ref{eq:lasso}) for two datasets $X_1$ and $X_2$
through a new regularization term for the variation between two parameters $\pnorm{\Vec{\theta}_1 - \Vec{\theta}_2}{1}$, 
\begin{align}
	\min_{ \Vec{\theta}_1, \Vec{\theta}_2 } \sum_{i=1}^2 \left\{ \frac{1}{2} \pnorm{X_{i, j} - X_{i, \setminus j} \Vec{\theta}_i}{2}^2 
		+ \rho \pnorm{\Vec{\theta}_i}{1} \right\} + \gm \pnorm{\Vec{\theta}_1 - \Vec{\theta}_2}{1} \, ,
	\label{eq:flasso}
\end{align}
where $\gamma \geq 0$ is a regularization parameter for the variation.
The new term enforces the variation of some elements in two parameters to shrink to zeros.
They also provided a coordinate descent-based optimization procedure for the above problem.

%==============================================================================================
\section{Learning Common Patterns in Multiple GGMs}
\label{sec:cssl}

The preceding work by~\cite{zhanglearning} adopted the idea of the Fused-Lasso type technique 
using the specific formulation of the two datasets situation.
In our study, we introduce a new framework, a {\it Common Substructure Learning} (CSSL), 
for finding invariant patterns in multiple dependency structures that is applicable to the general case of $N$ datasets.

%==============================================================================================
\subsection{Common Substructure Learning Problem}

We first formalize what invariance we are aiming to detect in multiple dependency structures.
To begin with, we assume that the number of variables in each dataset is the same, so they are all $d$-dimensional.
Also, the identities of each variable are the same.
For instance, $x_1$ is always a value from the same sensor while its behavior may change across datasets.
We then define a common substructure for multiple GGMs as follows.
\begin{definition}[Common Substructure of Multiple GGMs]
	\label{def:css}
	Let $\Lm_1,$ $\Lm_2,$ $\ldots,$ $\Lm_N$ be precision matrices corresponding to each GGM.
	Then, the common substructure of the GGMs is expressed by an adjacency matrix $\Th \in \R^{d \times d}$ defined as
	\begin{eqnarray}
		\Th_{jj'} = \left\{
			\begin{array}{lc}
				\Lm_{1, jj'} \; , & \; \text{if} 
					\;\; \Lm_{1, jj'} = \Lm_{2, jj'} = \ldots = \Lm_{N, jj'} \\
				0 \; , & \; \text{otherwise}
			\end{array}
			\right. \; .
			\label{eq:css}
	\end{eqnarray}
\end{definition}
Note this is a natural extension of the invariance notion adopted in the prior work by \cite{zhanglearning} for the case of two datasets.
With an ordinal sparsity assumption for GGMs, this definition leads the precision matrices to simultaneously have sparseness and commonness. 
That is:
\begin{itemize}
	\item Sparseness: $\Lm_{i, jj'} = 0$ for some $1 \leq i \leq N$ and $1 \leq j, j' \leq d$, 
	\item Commonness: $\Lm_{1, jj'} = \Lm_{2, jj'} = \ldots = \Lm_{N, jj'}$ for some $1 \leq j, j' \leq d$.
\end{itemize}

Under the above commonness, the basic idea of our framework is to parametrize each precision matrix $\Lm_i$
using two components, a common substructure $\Th$ and an individual substructure $\Om_i \in \R^{d \times d}$:
\begin{align}
	\Lm_i = \Th + \Om_i \, .
	\label{eq:prec_fact}
\end{align}
Here, each individual substructure matrix $\Om_i$ is composed of non-zero entries that are
not common across the $N$ precision matrices.

In the preceding formulation~(\ref{eq:flasso}), some entries in the two precision matrices are shrunk to the same value 
owing to the effect of the term $\pnorm{\Vec{\theta}_1 - \Vec{\theta}_2}{1}$.
In the proposed parameterization, such commonness corresponds to the case when some entries of the individual substructures are simultaneously zero, 
that is, $\Om_{1, jj'} = \Om_{2, jj'} = \ldots = \Om_{N, jj'} = 0$.
Hence, the non-zero common value is expressed by a common substructure matrix $\Th$.
These facts motivate us to regularize the individual substructures through the grouped regularization $\pnorm{\Om}{1, p}$.
On the other hand, we expect a common substructure $\Th$ to be sparse so that we can interpret it easily.
To that end, we adopt an ordinary $\ell_1$-regularization $\pnorm{\Th}{1}$ and the overall problem is summarized as follows:
\begin{align}
	& \max_{ \Th, \{ \Om_i \}_{i=1}^N } \sum_{i=1}^N t_i \ell(\Th + \Om_i ; S_i) - \rho \pnorm{\Th}{1} - \gm \pnorm{\Om}{1, p} \nonumber \\
	& \hspace{8pt} \st \Th + \Om_i \in \PD \;\; (1 \leq i \leq N) \, ,
	\label{eq:cssl}
\end{align}
with regularization parameters $\rho, \gm \geq 0$.
Since $-\ell(\Th + \Om_i; S_i), \|\Th\|_1$ and $\|\Om\|_{1, p}$ are all convex, the entire formulation is again a convex optimization problem.
We refer to this problem as {\it Common Substructure Learning} (CSSL).
Note that in the above formulation, we have slightly relaxed the condition of commonness to allow 
$\Th_{jj'}$ and $\Om_{i, jj'}$ to become simultaneously non-zeros which is contrary to Definition~(\ref{eq:css}).
We correct this point by applying the criterion~(\ref{eq:css}) to the resulting precision matrices $\hat{\Lm}_1, \hat{\Lm}_2, \ldots, \hat{\Lm}_N$
in the post processing stage to extract only truly common entries.

Here, we list two important properties of the CSSL problem (\ref{eq:cssl}), a dual problem and the bound on eigenvalues.
We first present the dual problem, which plays an important role in constructing an efficient optimization algorithm in the next section.
\begin{proposition}[Dual of CSSL]
	\label{prop:cssl_dual}
	The dual problem of CSSL~(\ref{eq:cssl}) is 
	\begin{align}
		& \min_{ \{ W_i ; W_i \in \PD \}_{i=1}^N } - \sum_{i=1}^N t_i \log \det W_i - d \, , \nonumber \\
		& \hspace{8pt} \st \left| \sum_{i=1}^N t_i \left( W_{i, jj'} - S_{i, jj'} \right) \right| \leq \rho \; , \nonumber \\
		& \hspace{24pt} \left( \sum_{i=1}^N t_i^q |W_{i, jj'} - S_{i, jj'}|^q \right)^\frac{1}{q} \leq \gm \; \; (1 \leq j, j' \leq d) \, ,
		\label{eq:cssl_dual}
	\end{align}
	where $q$ is a parameter satisfying $p^{-1} + q^{-1} = 1$.
	The resulting matrices of the dual problem $W_i^*$ are related to the optimal precision matrices $\Lm_i^*$ 
	through the inverse, $\Lm_i^* = {W_i^*}^{-1}$.
\end{proposition}

In both the primal and dual formulations~(\ref{eq:cssl}), (\ref{eq:cssl_dual}), we enforced the positive definiteness constraints, 
$\Lm_i = \Th + \Om_i \in \PD$ and $W_i \in \PD$ so that the matrices are valid precision or covariance matrices.
Here, we show that they can be tightened according to the next theorem.
\begin{theorem}[Bounds on Eigenvalues]
	\label{th:eig_bound}
	The optimal precision matrices for the CSSL~(\ref{eq:cssl}) $\Lm_1^*, \Lm_2^*, \ldots, \Lm_N^*$ with $0 < \rho < N^\frac{1}{p} \gm < \infty$
	have bounded eigenvalues $\lambda_i^{\min} I_d \preceq \Lm_i^* \preceq \lambda_i^{\max}I_d$, where the bounding parameters 
	$\lambda_i^{\min}$ and $\lambda_i^{\max}$ are 
	\begin{align*}
		\lambda_i^{\min} = \frac{t_i}{t_i \snorm{S_i} + d \gm} \; , \;\; \lambda_i^{\max} = \frac{N^\frac{1}{p}d^2}{\rho} \, .
	\end{align*}
\end{theorem}
Using this result, we can replace the constraint $\Lm_i \in \PD$ with the tighter
$\Lm_i \in \tilde{\PD}_i = \{ A \in \R^{d \times d} ; A \succeq \lambda_i^{\min} I_d \}$, 
and similarly $W_i \in \{ A \in \R^{d \times d} ; A \succeq {\lambda_i^{\max}}^{-1} I_d \}$.
Note that this update is practically important when constructing an optimization algorithm.
Since the new constraint set $\tilde{\PD}_i$ is closed, we can project points out of the constraint set onto the boundary, 
which is unavailable for the original open set $\PD$.

%==============================================================================================
\subsection{Interpretations of CSSL}
\label{sec:interpret_cssl}

The proposed CSSL problem~(\ref{eq:cssl}) can be interpreted as a generalization of 
an ordinary SICS problem~(\ref{eq:sics}) and its multi-task extension MSICS~(\ref{eq:msics}).
In the case that $\gm \rightarrow \infty$, the solution to the CSSL is $\Om_1 = \Om_2 = \ldots = \Om_N = 0_{d \times d}$, 
which means that all precision matrices are equal and are represented by a single matrix $\Th$.
Such $\Th$ is available by solving the SICS problem~(\ref{eq:sics}) with $S = \sum_{i=1}^N t_i S_i$.
On the other hand, if $\rho \geq N^\frac{1}{p} \gm$, the common substructure $\Th$ becomes zero.
This fact follows from the relationship between the $\ell_p$-norms:
\begin{align*}
	\hspace{-8pt}
	\gm \pnorm{\Th + \Om_i}{1, p} \leq N^\frac{1}{p} \gm \pnorm{\Th}{1} + \gm \pnorm{\Om}{1, p}
		\leq \rho \pnorm{\Th}{1} + \gm \pnorm{\Om}{1, p} \, .
\end{align*}
Suppose that the common substructure is non-zero, that is, $\Th \neq 0_{d \times d}$, then the above inequality means
that the update $\Om_i \leftarrow \Th + \Om_i$ and $\Th \leftarrow 0_{d \times d}$ improves the objective function value (\ref{eq:cssl})
without changing the resulting precision matrix $\Lm_i = \Th + \Om_i$, and thus the solution must be $\Th = 0_{d \times d}$.
Under this situation, the CSSL problem~(\ref{eq:cssl}) coincides with MSICS~(\ref{eq:msics}).
For the proper parameters $\rho < N^\frac{1}{p} \gm < \infty$, the CSSL problem~(\ref{eq:cssl}) is the intermediate of those two problems.

The CSSL problem can also be interpreted from a distributional perspective.
From the relationship between the Lagrangian expression and the constrained optimization problem, the CSSL problem (\ref{eq:cssl}) 
is equivalent to solving a set of $N$ maximum likelihood estimation problems~(\ref{eq:ml}) under the additional constraints
\begin{align}
	\pnorm{\Th}{1} \leq \eta \; , \; \pnorm{\Om}{1, p} \leq \eta' \, ,
	\label{eq:norm_bound}
\end{align}
for some properly chosen positive constants $\eta, \eta'$.
Moreover, we have
\begin{align*}
	\max_{1 \leq i < i' \leq N} \pnorm{\Om_i - \Om_{i'}}{1}
		& \leq \max_{1 \leq i < i' \leq N} \sum_{j, j' = 1}^d \left( \left| \Om_{i, jj'} \right| + \left| \Om_{i', jj'} \right| \right) \\
	& \leq 2 \pnorm{\Om}{1, \infty} \leq 2 \pnorm{\Om}{1, p} \, ,
\end{align*}
where the second inequality comes from the fact that exchanging the order of $\max_{1 \leq i < i' \leq N}$ and $\sum_{j, j'=1}^d$ 
produces the upper bound.
The last inequality is an ordinary relationship between $\ell_p$-norms.
These relations and the fact that $\Lm_i - \Lm_{i'} = \Om_i - \Om_{i'}$ lead to the bound
\begin{align*}
	\max_{1 \leq i < i' \leq N} \pnorm{\Lm_i - \Lm_{i'}}{1} \leq 2 \eta' \, .
\end{align*}
Hence, from the result of \citet[Lemma~23]{honoriolipschitz} and general matrix norm rules, 
the left-hand side of this inequality can be interpreted as the upper bound of the KL divergence between two distributions 
$p_i(\Vec{x}) = \ND(\Vec{0}_d, \Lm_i^{-1})$ and $p_{i'}(\Vec{x}) = \ND(\Vec{0}_d, \Lm_{i'}^{-1})$.
With these properties, we can interpret the second constraint in (\ref{eq:norm_bound}) as a constraint on the similarity among distributions: 
\begin{align*}
	\max_{1 \leq i, i' \leq N} \KL (p_i(\Vec{x}) || p_{i'}(\Vec{x})) \leq 2 \eta' \max_{1 \leq i \leq N} \| \Lm_i^{-1} \|_{\rm S} \, ,
\end{align*}
where $D_{\text KL}(p_i(\Vec{x}) || p_{i'}(\Vec{x}))$ denotes a KL divergence between two distributions $p_i(\Vec{x})$ and $p_{i'}(\Vec{x})$.
From Theorem~\ref{th:eig_bound}, the optimal parameters $\Lm_1^*, \Lm_2^*, \ldots, \Lm_N^*$ have bounded spectral norms for a finite $\gm$, 
and thus this upper bound on the KL divergence is always valid.
Moreover, we can further extend this bound into the extreme case $\gm \rightarrow \infty$ and $\eta' \rightarrow 0$.
%Moreover, we can verify that this bound is valid even in the extreme $\gm \rightarrow \infty$ and $\eta' \rightarrow 0$.
%As we have discussed before, this corresponds to the case $\Om_1 = \Om_2 = \ldots = \Om_N = 0_{d \times d}$ 
As we have discussed before, this is the case $\Om_1 = \Om_2 = \ldots = \Om_N = 0_{d \times d}$ 
and the problem is equivalent to solving a single SICS problem for $\Th$ with $S = \sum_{i=1}^N S_i$.
Hence, from \citet[Theorem~1]{banerjee2008model}, we can see that the resulting precision matrices still have finite eigenvalues for $\rho > 0$, 
and the right hand side of the above inequality goes to zero.
This means that the resulting distributions represented by precision matrices derived from CSSL~(\ref{eq:cssl}) 
have to be similar to one another at some level and they can be even identical in the extreme case.
Note that MSICS~(\ref{eq:msics}) is a special case of CSSL when $\Th = 0_{d \times d}$ and thus the same upper bound holds, 
although there is the significant distinction that the parameter $\eta'$ in MSICS~(\ref{eq:msics}) also affects 
the sparsity of the resulting precision matrices while CSSL~(\ref{eq:cssl}) can control the sparsity through the other hyper-parameter $\rho$. 

%==============================================================================================
\subsection{Connection to Additive Sparsity Models}

In this section, we discuss some connections of the CSSL problem (\ref{eq:cssl}) to {\it Additive Sparsity Models}
\citep{jalali2010dirty,chandrasekaran2010latent,agarwal2011noisy,candes2011robust,obozinski2011group}.
In general additive sparsity models, the objective parameter we want to estimate is modeled as the sum of two components, as in (\ref{eq:prec_fact}).
Hence, these two parameters are estimated using sparsity inducing norms such as an $\ell_1$-norm and a trace-norm.
In this sense, CSSL can be seen as a specific example of additive sparsity models where we use
the combination of an $\ell_1$-regularization and a group-wise regularization.

Here, we point out two close works from \citet{jalali2010dirty} and \citet{chandrasekaran2010latent}.
The former considers the multi-task least squares regression problem under the combination of $\ell_1$, group-wise regularizations.
Their basic idea is quite close to ours in that some regression parameters can be close to each other across datasets.
They also prove the advantage of combining two regularizations over using only one theoretically and numerically.
The latter study is on GGMs but with different sparsity assumptions from ours.
They show that the additive sparsity model naturally appears in GGM when there are latent variables.
In such a situation, the first component in the additive sparsity model corresponds to the precision matrix between observed variables
while the latter component is an interaction between latent variables.
This insight is also available for interpreting our model (\ref{eq:prec_fact}), that is, 
a common interaction among observed variables is contaminated by the effect of latent variables which are different for each dataset.

%==============================================================================================
\section{Optimization via DAL-ADMM}
\label{sec:alg}

In this section, we present the optimization algorithm for solving the CSSL problem (\ref{eq:cssl_dual}).
Our basic approach here is to adopt the Augmented Lagrangian techniques~\citep{hestenes1969multiplier,powell1967method}.
In a prior study, \citet{tomioka2011super} have shown that solving a dual problem using the Augmented Lagrangian, 
which is referred to as {\it Dual Augmented Lagrangian} (DAL), is preferable for the case when the primal loss is badly conditioned. 
See \citet[Table~3]{tomioka2011super} and the discussion therein.
This is actually the case we are faced with, as summarized in the next theorem.
\begin{theorem}
	\label{th:cssl_cond}
	The Hessian matrix of the CSSL primal loss function $\sum_{i=1}^N t_i \ell(\Th + \Om_i; S_i)$ is rank-deficient 
	while the Hessian matrix of the CSSL dual loss function $- \sum_{i=1}^N t_i \log \det W_i$ is always full rank
	for $0 < \rho < N^\frac{1}{p} \gm < \infty$.
\end{theorem}
This fact motivates us to solve the dual problem rather than the primal problem.
To that end, we construct an algorithm based on the DAL approach.

%==============================================================================================
\subsection{DAL-ADMM Algorithm}

The basic structure of the proposed algorithm is based on the idea of DAL.
However, while the original DAL requires solving the inner problem almost exactly~\citep{tomioka2011super}, 
we take an alternative approach using ADMM~\citep{gabay1976dual,boyd2011distributed} that makes the entire procedure dramatically simple.

To begin with, we rewrite the CSSL dual problem (\ref{eq:cssl_dual}) in the following equivalent form: 
\begin{align}
	& \min_{ \{ W_i, Y_i; W_i \in \PD \}_{i=1}^N } - \sum_{i=1}^N t_i \log \det W_i \nonumber \\
	& \hspace{8pt} \st t_i W_i - Y_i - t_i S_i = 0 \;\; (1 \leq i \leq N) \; , \nonumber \\ 
	& \hspace{24pt} \left| \sum_{i=1}^N Y_{i, jj'} \right| \leq \rho \, , \;
		\left( \sum_{i=1}^N \left| Y_{i, jj'} \right|^q \right)^\frac{1}{q} \leq \gm \;\; (1 \leq j, j' \leq d) \, .
	\label{eq:cssl_dual3}
\end{align}
Based on this expression, we define the following Augmented Lagrangian function: 
\begin{align}
	\hspace{-16pt}
	\Lag_\beta (W, Y, Z) =& - \sum_{i=1}^N t_i \log \det W_i + \delta_\rho(Y) + \tilde{\delta}_{\gm}^q(Y) \nonumber \\
	& + \tr{Z^\top \left( T W - Y - T \Sigma \right)} + \frac{\beta}{2} \pnorm{T W - Y - T \Sigma}{2}^2 \, ,
	\label{eq:dal}
\end{align}
where $\beta$ is a nonnegative parameter and $\Sigma$, $W$, $Y$ and $Z$ are the concatenated matrices 
$\Sigma = \matrix{cccc}{S_1 & S_2 & \ldots & S_N}^\top$, $W = \matrix{cccc}{W_1 & W_2 & \ldots & W_N}^\top$, 
$Y = \matrix{cccc}{Y_1 & Y_2 & \ldots & Y_N}^\top$ and $Z = \matrix{cccc}{Z_1 & Z_2 & \ldots & Z_N}^\top$, 
and $T$ is as the matrix $T = \diag{[t_1, t_2, \ldots, t_N]^\top} \otimes I_d$, where $\otimes$ denotes the Kronecker product
and $I_d$ is the $d$-dimensional identity matrix.
We also defined the functions $\delta_\rho(Y)$ and $\tilde{\delta}_\gm^q(Y)$ as
\begin{align*}
	& \hspace{-10pt} \delta_\rho(Y) = \left\{\begin{array}{cl}
		0 \, , & \text{if} \; \left|\sum_{i=1}^N Y_{i, jj'} \right| \leq \rho \;\; \text{for} \;\; 1 \leq j, j' \leq d \\
		\infty \, , & \text{otherwise}
	\end{array}\right. , \\
	& \hspace{-10pt} \tilde{\delta}_{\gm}^q(Y) = \left\{\begin{array}{cl}
		0 \, , & \text{if} \; \left( \sum_{i=1}^N \left|Y_{i, jj'}\right|^q \right)^\frac{1}{q} \leq \gm \;\;
			\text{for} \;\; 1 \leq j, j' \leq d \\
		\infty \, , & \text{otherwise}
	\end{array}\right. .
\end{align*}
In the Augmented Lagrangian function (\ref{eq:dal}), the optimal precision matrix $\Lm_i^*$ is represented by the optimal dual variable $Z_i^*$.
This can be verified through a simple calculation.
We set the derivative of the unaugmented Lagrangian $\Lag_0(W, Y, Z)$ over $W_i$ to zeros and find that
\begin{align*}
	{W_i^*}^{-1} = Z_i^* \, ,
\end{align*}
which implies that $\Lm_i^* = Z_i^*$ from Proposition \ref{prop:cssl_dual}.
This follows since the solution to (\ref{eq:cssl_dual3}) must be the saddle point of the unaugmented Lagrangian function $\Lag_0(W, Y, Z)$.

We solve problem (\ref{eq:cssl_dual3}) using ADMM by iteratively applying the following three steps until convergence: 
\begin{align*}
	\left\{\begin{array}{cl}
		W^{(k+1)} & \!\! \in \argmin{ \{ W_i; W_i \in \PD \}_{i=1}^N } \; \Lag_\beta (W, Y^{(k)}, Z^{(k)}) \\
		Y^{(k+1)} & \!\! \in \argmin{ Y } \; \Lag_\beta (W^{(k+1)}, Y, Z^{(k)}) \\
		Z^{(k+1)} & \!\! = Z^{(k)} + \beta \left( T W^{(k+1)} - Y^{(k+1)} - T \Sigma \right)
	\end{array}\right. \, .
\end{align*}
Hence, using ADMM, convergence of the dual variable $Z$ to the optimal parameter $Z^*$ is guaranteed 
as the number of iterations tends to infinity~\citep[Section~3.2]{boyd2011distributed}.
This means we can find the optimal precision matrices $\Lm_1^*, \Lm_2^*, \ldots, \Lm_N^*$ using DAL-ADMM.
In the following two subsections, we give the update procedures for $W$ and $Y$.

%==============================================================================================
\subsection{Inner Optimization Problem: Update of $W$}

The update of $W$ can be factorized into $N$ independent problems where each problem defines an update of $W_i$: 
\begin{align*}
	\hspace{-10pt}
	\min_{ W_i \in \PD } \; - t_i \log \det W_i + t_i \tr{{Z_i^{(k)}}^\top W_i} 
		+ \frac{\beta}{2} \pnorm{t_i W_i - Y_i^{(k)} - t_i S_i}{2}^2 \, .
\end{align*}
By setting the derivative over $W_i$ to zero, we obtain
\begin{align*}
	W_i - \left( \frac{1}{t_i}Y_i^{(k)} - \frac{1}{\beta t_i }Z_i^{(k)} + S_i \right) - \frac{1}{\beta t_i}W_i^{-1} = 0_{d \times d} \, .
\end{align*}
Now, write the eigen-decomposition as $\frac{1}{t_i}Y_i^{(k)} - \frac{1}{\beta t_i}Z_i^{(k)} + S_i = PDP^\top$
with $D = \diag{\sg_1, \sg_2, \ldots, \sg_d}$ and $P^\top P = PP^\top = I_d$.
Then, the above matrix equation has a solution of the form $W_i = P\tilde{D}P^\top$ with $\tilde{D} = \diag{\tsg_1, \tsg_2, \ldots, \tsg_d}$.
The equation for each eigenvalue is $\tsg_m - \sg_m - \frac{1}{\beta t_i}\tsg_m^{-1} = 0 \; (1 \leq m \leq d)$, which has the analytic solution
\begin{align*}
	\tsg_m = \frac{\sg_m + \sqrt{\sg_m^2 + \frac{4}{\beta t_i}}}{2} \, .
\end{align*}
Note the positive definiteness of $W_i$ is automatically fulfilled since $\tsg_m > 0$ for $\beta > 0$.

%==============================================================================================
\subsection{Inner Optimization Problem: Update of $Y$}

\begin{table*}[t]
\centering
\caption{Solutions to problem~(\ref{eq:yupdate2}) for $q = 1, 2$ and $\infty$:
see the corresponding appendix for further details.
An operator $T_\gm (*)$ in $\Vec{y} \in \pC_2$ for $q = \infty$ is a thresholding for each $y_{0, i}$, 
that is, $y_i = \sgn{y_{0, i}} \min(|y_{0, i}|, \gm)$.}
\label{tab:yupdate}
\small
\begin{tabular}{|c||c|c|c|}
	\hline
		& \begin{minipage}{0.25\textwidth} \begin{center} $q = 1$ \end{center} \end{minipage} 
		& \begin{minipage}{0.25\textwidth} \begin{center} $q = 2$ \end{center} \end{minipage} 
		& \begin{minipage}{0.25\textwidth} \begin{center} $q = \infty$ \end{center} \end{minipage} \\
	\hline
	\multirow{2}{*}{$\Vec{y}_0 \in \C$} 
		& \multicolumn{3}{|c|}{\multirow{2}{*}{$\Vec{y} = \Vec{y}_0$}} \\
	& \multicolumn{3}{c|}{} \\
	\hline
	\multirow{3}{*}{$\Vec{y} \in \pC_1$}
		& \multicolumn{3}{|c|}
			{\multirow{3}{*}{
			\shortstack{$\displaystyle \Vec{y} 
				= \Vec{y}_0 - \frac{\Vec{1}_N^\top \Vec{y}_0 - \rho \, \sgn{\Vec{1}_N^\top \Vec{y}_0}}{N}\Vec{1}_N$ 
				\enspace (\ref{app:c1})}}} \\
	& \multicolumn{3}{c|}{} \\
	& \multicolumn{3}{c|}{} \\
	\hline
	\multirow{3}{*}{$\Vec{y} \in \pC_2$} 
		& \multirow{3}{*}{\shortstack{Continuous Quadratic \\Knapsack Problem \\(\ref{app:c2_1})}} 
		& \multirow{3}{*}{\shortstack{$\displaystyle \Vec{y} = \frac{\gm}{\pnorm{\Vec{y}_0}{2}}\Vec{y}_0$ \\(\ref{app:c2_2})}} 
		& \multirow{3}{*}{\shortstack{$\Vec{y} = T_\gm (\Vec{y}_0)$ \\(\ref{app:c2_inf})}} \\
	& & & \\
	& & & \\
	\hline
	\multirow{3}{*}{$\Vec{y} \in \pC_3$} 
		& \multirow{3}{*}{\shortstack{Continuous Quadratic \\ Knapsack Problem \\(\ref{app:c3_1})}} 
		& \multirow{3}{*}{\shortstack{Analytic Solution \\(\ref{app:c3_2})}}
		& \multirow{3}{*}{\shortstack{Continuous Quadratic \\ Knapsack Problem \\(\ref{app:c3_inf})}} \\
	& & & \\
	& & & \\
	\hline
\end{tabular}
\vspace{-2.5ex}
\end{table*}

The update of $Y$ is formulated as 
\begin{align*}
	\hspace{-6pt}
	\min_Y \; \delta_\rho(Y) + \tilde{\delta}_\gm^q(Y) - \tr{{Z^{(k)}}^\top Y}
		+ \frac{\beta}{2} \pnorm{T W^{(k + 1)} - Y - T \Sigma}{2}^2 \, ,
\end{align*}
or equivalently, the projection $Y = \proj{Y_0}{\mathcal{A}}$ of $Y_0 = T W^{(k + 1)} + \frac{1}{\beta}Z^{(k)} - T \Sigma$ onto the set
$\mathcal{A} = \left\{ Y = \matrix{cccc}{Y_1 & Y_2 & \ldots & Y_N}^\top ; \right.$
$\left| \sum_{i=1}^N Y_{i, jj'} \right| \leq \rho \; ,$
$\left. \left( \sum_{i=1}^N \left| Y_{i, jj'} \right|^q \right)^\frac{1}{q} \leq \gm \; , \forall j, j' \right\}$,
where $\text{proj}(*, *)$ is a projection function defined as
\begin{align*}
	\proj{V}{\mathcal{B}} = \argmin{U \in \mathcal{B}} \frac{1}{2} \pnorm{U - V}{2}^2 \, .
\end{align*}
We can further decompose this problem into $\Order(d^2)$ problems over $\Vec{y} = (Y_{1, jj'}, Y_{2, jj'}, \ldots, Y_{N, jj'})^\top$ 
for each $(j, j')$th entry.
Hence, each problem is
\begin{align}
	\Vec{y} = \proj{\Vec{y}_0}{\C} \, ,
	\label{eq:yupdate2}
\end{align}
where $\Vec{y}_0$ is an $N$-dimensional vector with the $i$th component equal to
$y_{0, i} = t_i W_{i, jj'}^{(k+1)} + \frac{1}{\beta}Z_{i, jj'}^{(k)} - t_i S_{i, jj'}$, 
and where the constraint set is $\C = \{ \Vec{u} \in \R^N ; |\Vec{1}_N^\top \Vec{u}| \leq \rho, \, \pnorm{\Vec{u}}{q} \leq \gm \}$
with $\Vec{1}_N$ being an $N$-dimensional vector of ones.

For any $q \in [1, \infty]$, problem~(\ref{eq:yupdate2}) has a trivial solution $\Vec{y} = \Vec{y}_0$ if $\Vec{y}_0 \in \C$.
In the remaining cases, that is, $|\Vec{1}_N^\top \Vec{y}_0| > \rho$ or $\pnorm{\Vec{y}_0}{q} > \gm$, 
the solution is on the boundary of the constraint set 
%$\pC = \{ \Vec{u} ; |\Vec{1}_N^\top \Vec{u}| = \rho, \pnorm{\Vec{u}}{q} \leq \gm \} \cup \{ \Vec{u} ; |\Vec{1}_N^\top \Vec{u}| \leq \rho, \pnorm{\Vec{u}}{q} = \gm \}$ 
$\pC = \{ \Vec{u} ; |\Vec{1}_N^\top \Vec{u}| = \rho, \pnorm{\Vec{u}}{q} \leq \gm \} \cap \{ \Vec{u} ; |\Vec{1}_N^\top \Vec{u}| \leq \rho, \pnorm{\Vec{u}}{q} = \gm \}$ 
owing to the convexity of the objective function.
Thus, the problem can be reduced to a search of the boundary.
However, even though the constraint set $\C$ is convex, it is an intersection of two sets and the shape of the boundary $\pC$ is rather complicated.
Therefore, we do not search the boundary $\pC$ directly, but solve a set of simpler problems instead.
The basic approach is to classify the boundary into three parts, 
$\pC_1 = \{ \Vec{u} ; |\Vec{1}_N^\top \Vec{u}| = \rho, \, \pnorm{\Vec{u}}{q} \neq \gm \}$, 
$\pC_2 = \{ \Vec{u} ; |\Vec{1}_N^\top \Vec{u}| \neq \rho, \, \pnorm{\Vec{u}}{q} = \gm \}$
and $\pC_3 = \{ \Vec{u} ; |\Vec{1}_N^\top \Vec{u}| = \rho, \, \pnorm{\Vec{u}}{q} = \gm \}$.
The problems we solve here are modified versions of~(\ref{eq:yupdate2}), replacing the constraint with $\Vec{y} \in \pC_m$ for each $m \in \{1, 2, 3\}$: 
\begin{align}
	\Vec{y} = \proj{\Vec{y}_0}{\pC_m} \, .
	\label{eq:yupdate_sub}
\end{align}
Note that $\pC_1$ and $\pC_2$ involve infeasible solutions to the problem~(\ref{eq:yupdate2}).
For example, a point $\Vec{y}$ with $\pnorm{\Vec{y}}{q} > \gm$ is infeasible even if $\Vec{y} \in \pC_1$, 
while these three regions covers the entire boundary of the constraint set $\pC \subset \cup_{m=1}^3 \pC_m$.
This guarantees that we can search the entire boundary $\partial \mathcal{C}$ indirectly by searching the sets $\pC_m \, (m = 1, 2, 3)$ instead.
Hence, if neither of the solutions to (\ref{eq:yupdate_sub}) for $\Vec{y} \in \pC_1$ and $\Vec{y} \in \pC_2$
are involved in $\C$, the solution to (\ref{eq:yupdate2}) is in $\pC_3$.
We can take advantage of this property to construct an efficient solution procedure.
We first solve problems~(\ref{eq:yupdate_sub}) for $\Vec{y} \in \pC_1$ and $\Vec{y} \in \pC_2$, respectively, 
and if neither of solutions is in $\C$, then we solve (\ref{eq:yupdate_sub}) for $\Vec{y} \in \pC_3$.
In this paper, we focus on the specific cases $q = 1, 2$ and $\infty$, since efficient solution procedures are available.
In \tablename~\ref{tab:yupdate}, we summarized the solutions to problem~(\ref{eq:yupdate2}).
For further details, see~\ref{app:yupdate}.

%==============================================================================================
\subsection{Convergence Criteria}

Although the asymptotic convergence of $Z^{(k)}$ as $k \rightarrow \infty$ is theoretically guaranteed, 
in practice we need to stop the iteration at some point.
A major stopping criterion is the duality-gap, the difference between the primal and dual objective function values.
Let $f(W)$ be the objective function in (\ref{eq:cssl_dual}) and let $g(\Th, \Om)$ be the one in (\ref{eq:cssl}).
Then the duality-gap at the $k$th iteration is defined as
\begin{align*}
	\text{duality-gap} = f(\tilde{W}^{(k)}) - \max_{1 \leq k' \leq k} g(\tilde{\Th}^{(k')}, \tilde{\Om}^{(k')}) \, ,
\end{align*}
where $\tilde{W}^{(k)}$, $\tilde{\Th}^{(k)}$ and $\tilde{\Om}^{(k)}$ denote parameters estimated in the $k$th step
after proper projections and transformations.
We need these modifications of variables since the estimators in intermediate steps are not necessarily feasible.
For example, $W^{(k)}$ does not need to satisfy the constraints in (\ref{eq:cssl_dual})
since they are imposed only on a variable $Y$ in the DAL-ADMM setting~(\ref{eq:cssl_dual3}).
The projected variable $\tilde{W}^{(k)}$ is $\tilde{W}^{(k)} = T^{-1} \tilde{Y}^{(k)} + \Sigma$ 
where $\tilde{Y}^{(k)} = \proj{Y_0^{(k)}}{\mathcal{A}}$ and $Y_0^{(k)} = T (W^{(k)} - \Sigma)$.
The same goes for $\Lm^{(k)} = Z^{(k)}$.
An estimator $\Lm_i^{(k)}$ is not necessarily positive definite, and thus we project them as $\tilde{\Lm}_i^{(k)} = \proj{\Lm_i^{(k)}}{\tilde{\PD}_i}$.
This projection is available in the following manner.
Let $\Lm_i^{(k)} = PDP^\top$ be an eigen-decomposition with a diagonal matrix $D = \text{diag}(\sigma_1, \sigma_2, \ldots, \sigma_d)$. 
Then the projected matrix is $\tilde{\Lm}_i^{(k)} = P\tilde{D}P^\top$, 
where each element of $\tilde{D} = \text{diag}(\tilde{\sigma}_1, \tilde{\sigma}_2, \ldots, \tilde{\sigma}_d)$ 
is $\tilde{\sigma}_m = \max(\sigma_m, \lambda_i^{\min})$.
For computing the value of $g(\tilde{\Th}^{(k)}, \tilde{\Om}^{(k)})$, we need to further factorize $\tilde{\Lm}^{(k)}$
into $\tilde{\Th}^{(k)}$ and $\tilde{\Om}^{(k)}$.
This can be computed in an element-wise manner.
Let $\theta = \tilde{\Th}_{jj'}^{(k)}$, $\Om_{i, jj'}^{(k)} = \tilde{\Lm}_{i, jj'}^{(k)} - \theta$ 
and $\Vec{\lambda} = (\tilde{\Lm}_{1, jj'}^{(k)}, \tilde{\Lm}_{2, jj'}^{(k)}, \tilde{\Lm}_{N, jj'}^{(k)})^\top$. 
Then the problem we need to solve is
\begin{align*}
	\min_\theta \; \rho |\theta| + \gm \pnorm{\Vec{\lambda} - \theta \Vec{1}_N}{p} \, .
\end{align*}
For $p = 1$ and $\infty$, this function is piecewise linear with breakpoints $\{ 0, \lambda_1, \lambda_2, \ldots, \lambda_N \}$
and $\{ 0, \frac{\min_i \lambda_i + \max_{i'} \lambda_{i'}}{2} \}$, respectively.
Hence, the optimal $\theta$ is one of these breakpoints and can be found by searching the candidates.
For the case $p = 2$, the analytic solution is
\begin{align*}
	\hspace{-16pt}
	\theta = \frac{1}{N} \left\{ \Vec{1}_N^\top \tilde{\Vec{\lambda}} - \sgn{\Vec{1}_N^\top \tilde{\Vec{\lambda}}}
		\sqrt{ (\Vec{1}_N^\top \tilde{\Vec{\lambda}})^2 - N \frac{\gm^2 (\Vec{1}_N^\top \tilde{\Vec{\lambda}})^2 
		- \rho^2 \pnorm{\tilde{\Vec{\lambda}}}{2}^2}{\gm^2 N - \rho^2}} \right\} \, .
\end{align*}

Some other useful gaps are provided by \citet{boyd2011distributed}. 
The {\it primal-gap} measures how much the equality constraints in (\ref{eq:cssl_dual3}) is fulfilled, 
\begin{align*}
	\text{primal-gap} = \pnorm{T W^{(k)} - Y^{(k)} - T \Sigma}{2} \, ,
\end{align*}
while the {\it dual-gap} is a degree of the feasibility condition of the solution, defined as
\begin{align*}
	\text{dual-gap} = \beta \pnorm{T (Y^{(k + 1)} - Y^{(k)})}{2} \, .
\end{align*}

In our simulations in Sections~\ref{sec:simu} and~\ref{sec:anom}, we have evaluated both criteria.
We set two threshold parameters $\epsilon_{\rm gap}$ and $\epsilon_{\rm pdgap}$, and evaluated the conditions 
$\text{duality-gap} \leq \epsilon_{\rm gap}$ and $\max(\text{primal-gap}, \text{dual-gap}) \leq \epsilon_{\rm pdgap}$ in each iteration.
If one of two conditions is fulfilled, we regard the iteration as converged and output the result.
In the simulations in Sections~\ref{sec:simu} and~\ref{sec:anom}, we set $\epsilon_{\rm gap} = 10^{-5} d$ and $\epsilon_{\rm pdgap} = 10^{-5}$.

%==============================================================================================
\subsection{Computational Complexity}

In this section, we summarize the computational complexity of the proposed algorithm.
In the $W$ update step, the computational cost is dominated by the eigen-decomposition of a $d \times d$ matrix, 
which requires $\Order(d^3)$ operations, so the overall complexity is $\Order(Nd^3)$ for the update of $N$ matrices.
In the $Y$ update step, we need a projection $\proj{Y_0}{\mathcal{A}}$ which is divided into $\Order(d^2)$ subproblems.
For both $q=1$ and $q = \infty$, the most computationally expensive procedure is solving the continuous quadratic knapsack problem
which requires sorting $\Order(N)$ elements and has complexity $\Order(N \ln N)$~\footnote{See \ref{app:c2_1}, \ref{app:c3_1}, and \ref{app:c3_inf}.}.
In the case $q = 2$, the update is analytically available with $\Order(N)$ complexity.
The overall complexity for the $Y$ update is thus $\Order((N \ln N) d^2)$ for $q = 1, \infty$ and $\Order(Nd^2)$ for $q=2$.
The complexity for the $Z$ update is $\Order(N d^2)$.
In the convergence check, we need to calculate the projection $\text{proj}(\Lm_i^{(k)}, \tilde{\PD}_i)$
which has $\Order(d^3)$ complexity or $\Order(Nd^3)$ for $N$ matrices.
We also need the projection $\proj{Y_0^{(k)}}{\mathcal{A}}$ which is again $\Order((N \ln N)d^2)$ for $q = 1, \infty$ and $\Order(Nd^2)$ for $q = 2$.
Summarizing the above results, we conclude that the computational complexity of one update 
in DAL-ADMM is $\Order(Nd^3 + (N \ln N)d^2)$ for $q = 1, \infty$ and $\Order(Nd^3)$ for $q = 2$.
In many practical situations, the number of datasets $N$ is in the tens, while the dimensionality of the data $d$ can be a few hundred.
In such cases, $\ln N \ll d$ holds, and the entire complexity is approximately $\Order(Nd^3)$.
We note this is the least necessary complexity.
For an unregularized setting, the solution $\Lm_i^*$ is a maximum likelihood estimate $S_i^{-1}$, 
which requires $\Order(d^3)$ complexity for a matrix inverse and $\Order(Nd^3)$ for $N$ matrices.

Despite the theoretical complexity, the choice of $\beta$ is of practical importance since it affects the number of iterations needed until convergence.
We propose using the heuristic from \citet{boyd2011distributed}. 
In this heuristic, we update the value of $\beta = \beta^{(k)}$ in every steps following the next rule:
\begin{align*}
	\beta^{(k+1)} = \left\{\begin{array}{cl}
	2 \beta^{(k)} \, , & \text{if} \;\; \text{primal-gap} \geq 10 * \text{dual-gap} \\
	0.5 \beta^{(k)} \, , & \text{if} \;\; \text{dual-gap} \geq 10 * \text{primal-gap} \\
	\beta^{(k)} \, , & \text{otherwise}
	\end{array}\right. \, .
\end{align*}
While this does not give any theoretical guarantees on its performance, it does give us a pragmatic choice of $\beta$
and results in convergence with a smaller number of steps.

%==============================================================================================
\subsection{Heuristic Choice of Hyper--parameters}

In the CSSL problem~(\ref{eq:cssl}), the choice of hyper-parameters $\rho$ and $\gm$ affects the resulting precision matrices.
There are several approaches for choosing these, such as cross-validation~\citep{yuan2007model,guo2011joint} 
or the Bayesian information criterion~\citep{guo2011joint}.
Apart from selection techniques, the following result gives us some insight into $\rho$ and $\gm$, 
and is helpful for analyzing the data more intensively.

\begin{proposition}
	\label{prop:hyp_param}
	Let the bivariate common substructure $\Th$ and individual substructures $\Om_i$ be in the forms 
	$\Th = \matrix{cc}{0 & \theta \\ \theta & 0}$ and $\Om_i = \matrix{cc}{u_i & \omega_i \\ \omega_i & v_i}$,
	and consider the following CSSL problem with regularizations only on off-diagonal entries: 
	\begin{align}
		& \max_{ \Th, \{ \Om_i \}_{i=1}^N } \sum_{i=1}^N t_i \ell(\Th + \Om_i; S_i) - 2 \rho |\theta| - 2 \gm \pnorm{\Vec{\omega}}{p} \nonumber \\
		& \hspace{8pt} \st \Th + \Omega_i \in \PD \;\; (1 \leq i \leq N) \; ,
		\label{eq:cssl_bivar}
	\end{align}
	where $\Vec{\omega} = (\omega_1, \omega_2, \ldots, \omega_N)^\top$.
	Then the off-diagonal entries of the resulting precision matrices $\theta, \Vec{\omega}$ have the following property: 
	\begin{align*}
		\max_{1 \leq i \leq N} |r_i| \leq \gm \; \text{and} \; \left| \sum_{i=1}^N t_i r_i \right| \leq \rho
			\; \Rightarrow \;  \theta = 0, \; \Vec{\omega} = \Vec{0}_N \, ,
	\end{align*}
	where $r_i$ is the off-diagonal entry of $S_i$.
\end{proposition}

Although the result is specific to the bivariate case, we can use this as a guideline for choosing the hyper-parameters $\rho$ and $\gm$.
It also shows that $\rho$ and $\gm$ are not independent of each other, but rather they should change simultaneously proportional to 
$\max_{1 \leq i \leq N} |r_i|$ and $\left| \sum_{i=1}^N t_i r_i \right|$.
In particular, if each matrix $S_i$ is multiplied by some positive constant $c$, the above condition indicates 
that $\rho$ and $\gamma$ also need to be multiplied by $c$.
Such scale invariance is maintained only by a linear model between $\rho$ and $\gamma$.
Therefore, we construct the following heuristic based on this linear model.
\begin{enumerate}
	\item Assume that the linear relation $\left| \sum_{i=1}^N t_i S_{i, jj'} \right| = s_1 \max_{1 \leq i \leq N} |S_{i, jj'}| + s_0$
		holds for all entries $1 \leq j \leq j' \leq d$ for some $s_0, s_1 \in \R$.
	\item Estimate $s_0, s_1$ with least squares regression using the tuples
		$\left\{ \max_{1 \leq i \leq N} |S_{i, jj'}|, \left| \sum_{i=1}^N t_i S_{i, jj'} \right| \right\}_{1 \leq j \leq j' \leq d}$.
	\item Parameterize $\rho, \gm$ as $\rho = \max (s_1 \alpha + s_0, 0)$ and $\gm = \alpha$ using a parameter $\alpha$.
\end{enumerate}
This procedure provides an efficient way of tuning $\rho$ and $\gm$ simultaneously through a single parameter $\alpha$.

%==============================================================================================
\section{Simulation}
\label{sec:simu}

In this section, we investigate the performance of the proposed CSSL approach in finding common substructures 
among datasets through numerical simulations.

%==============================================================================================
\subsection{Generation of Synthetic Data}

We fist briefly summarize the data generation procedure for our simulations.
For the synthetic data, we need $N$ precision matrices with sparseness and commonness.
We tackle this problem in a two-stage approach.
We first generate a single sparse precision matrix, and then add some non-zero entries to make $N$ matrices 
where the additional patterns are individual to each other~\footnote{See \ref{app:gen_data} for further details.}. 
After $N$ precision matrices $\Lm_1, \Lm_2, \ldots, \Lm_N$ have been constructed, 
we generate $N$ datasets from the corresponding Gaussian distributions $\ND(\Vec{0}_d, \Lm_i^{-1})$ for $1 \leq i \leq N$.

%==============================================================================================
\subsection{Baseline Methods and Evaluation Measurements}

In the simulation, we adopt SICS~(\ref{eq:sics}) and MSICS~(\ref{eq:msics}) as baseline methods to compare with CSSL.
Since neither method is designed for finding a common substructure, we apply a heuristic to extract the substructure $\hat{\Th}$
from the estimated precision matrices $\hat{\Lm}_1, \hat{\Lm}_2, \ldots, \hat{\Lm}_N$. 
Note that, in SICS, each $\hat{\Lm}_i$ is estimated by solving~(\ref{eq:sics}) individually
while the set of matrices is estimated simultaneously in MSICS~(\ref{eq:msics}).
Following is the heuristic criterion used: 
\begin{align*}
	\hat{\Th}_{jj'} = \left\{\begin{array}{cl}
		\hat{\theta}_{jj'} \; , & \text{if} \; \max_{1 \leq i < i' \leq d} |\hat{\Lm}_{i, jj'} - \hat{\Lm}_{i', jj'}| \leq \epsilon \\
		0	\; , & \text{otherwise}
	\end{array}\right.
\end{align*}
where $\epsilon$ is some given threshold.
Here, to avoid selecting zero edges as parts of a common substructure, we set $\hat{\theta}_{jj'}$ to zero
if $\hat{\Lm}_{1, jj'} = \hat{\Lm}_{2, jj'} = \ldots = \hat{\Lm}_{N, jj'} = 0$ and one otherwise.
In our simulation, we select the threshold $\epsilon$ from the resulting precision matrices.
Specifically, we compute variations of estimators for each entry
$\left\{ \max_{1 \leq i < i' \leq N} |\hat{\Lm}_{i, jj'} - \hat{\Lm}_{i', jj'}| \right\}_{1 \leq j \leq j' \leq d}$, 
and then set $\epsilon$ as the $100\epsilon_0\%$ quantile.
This corresponds to considering the lower $100\epsilon_0\%$ varied entries as common.

In our simulation, we evaluate the common substructure detection performance through precision, recall and the F-measure.
While these values are defined based on the number of true positive, false positive and false negative detections, 
we slightly modify these measurements.
This is because finding common dependencies with higher amplitudes is much more important than
finding very small dependencies which can be approximated as zero in practice.
To that end, we adopt following weighted measurements, namely WTP (weighted true positive), WFP (weighted false positive), 
and WFN (weighted false negative), 
\begin{align*}
	\hspace{-8pt} \text{WTP} 
		& = \sum_{j < j'}^d \tilde{J}_{{\rm c}, jj'} \tilde{J}_{{\rm p}, jj'} J_{{\rm c}, jj'} \max_{1 \leq i \leq N} |\Lm_{i, jj'}| \, , \\
	\hspace{-8pt} \text{WFP} 
		& = \sum_{j < j'}^d \tilde{J}_{{\rm c}, jj'} \tilde{J}_{{\rm p}, jj'} (1 - J_{{\rm c}, jj'}) \max_{1 \leq i \leq N} |\Lm_{i, jj'}| \, , \\
	\hspace{-8pt} \text{WFN} 
		& = \sum_{j < j'}^d \left\{ \tilde{J}_{{\rm c}, jj'} (1 - \tilde{J}_{{\rm p}, jj'}) + (1 - \tilde{J}_{{\rm c}, jj'}) \right\} 
		J_{{\rm c}, jj'} \max_{1 \leq i \leq N} |\Lm_{i, jj'}| \, , 
\end{align*}
where $\tilde{J}_{{\rm c}, jj'}$, $\tilde{J}_{{\rm p}, jj'}$ and $J_{{\rm c}, jj'}$ are defined as
\begin{align*}
	& \tilde{J}_{{\rm c}, jj'} = I \left( \max_{1 \leq i < i' \leq N} |\hat{\Lm}_{i, jj'} - \hat{\Lm}_{i', jj'}| < \epsilon \right) \, , \\
	& \tilde{J}_{{\rm p}, jj'} = I \left( \max_{1 \leq i \leq N} |\hat{\Lm}_{i, jj'}| > 0 \right) \, , \\
	& J_{{\rm c}, jj'} = I \left( \max_{1 \leq i < i' \leq N} |\Lm_{i, jj'} - \Lm_{i', jj'}| = 0 \right) \, .
\end{align*}
Here, $I(P)$ is an indicator function that returns $1$ for a true statement $P$ and $0$ otherwise.
The modified measurements in the simulation are defined using these values as
\begin{align*}
	\text{Precision} & = \frac{\text{WTP}}{\text{WTP} + \text{WFP}} \, , \\
	\text{Recall} & = \frac{\text{WTP}}{\text{WTP} + \text{WFN}} \, , \\
	\text{F-measure} & = 2 \frac{\text{Precision} * \text{Recall}}{\text{Precision} + \text{Recall}} \, .
\end{align*}

In the simulation, we also observe whether the zero pattern in the precision matrices is properly recovered using each method.
We use the following F-measure for this evaluation, which we refer to the "${\rm F}_0$-measure" to distinguish it from the one above: 
\begin{align*}
	& \text{${\rm F}_0$-measure} = \frac{2 \text{TP}}{ 2 \text{TP} + \text{FP} + \text{FN}} \, , \\
	& \text{TP} = \sum_{i=1}^N \sum_{j < j'}^d I(\Lm_{i, jj'} = 0) I(\hat{\Lm}_{i, jj'} = 0) \, , \\
	& \text{FP} = \sum_{i=1}^N \sum_{j < j'}^d I(\Lm_{i, jj'} \neq 0) I(\hat{\Lm}_{i, jj'} = 0) \, , \\
	& \text{FN} = \sum_{i=1}^N \sum_{j < j'}^d I(\Lm_{i, jj'} = 0) I(\hat{\Lm}_{i, jj'} \neq 0) \, .
\end{align*}

%==============================================================================================
\subsection{Result}

\begin{table*}[t]
\centering
\caption{Simulation results for three cases ($d = 25, 50$ and $100$) with $N = 5$ datasets
evaluated by weighted precision, recall and F-measure, denoted by "Prec.", "Rec." and "F" in the table, respectively.
The "${\rm F}_0$" denotes the ${\rm F}_0$-measure for zero pattern identification.
Each simulation is conducted so that each dataset has $5d$ data points, and the measurements are averaged over $100$ random realization of datasets.
The numbers in brackets are standard deviations of each measurement.
Each of the three rows in SICS and MSICS corresponds to results for $\epsilon_0 = 0.5, 0.7$ and $0.9$ from the top.
We highlight the top three results for each measurement in bold font (except for "${\rm F}_0$").}
\label{tab:simu}
\small
\begin{tabular}{|c|c||ccccccc|}
	\hline
		& & CSSL & CSSL & CSSL & CSSL & \multirow{2}{*}{SICS} & MSICS & MSICS \\
		& & ($p=1$) & ($p=2$) & ($p=\infty$) & ($\gm=\infty$) & & ($p=2$) & ($p=\infty$) \\
	\hline
		& \multirow{3}{*}{\rotatebox{90}{Prec.}}
			& & & & & .14 (.14) & .38 (.21) & .54 (.23) \\
		& 	& \!\!{\bf .84 (.19)}\!\! & \!\!{\bf .70 (.16)}\!\! & \!\!{\bf .56 (.19)}\!\! & .48 (.20) 
					& .20 (.16) & .43 (.21) & .49 (.21) \\
		& 	& & & & & .33 (.16) & .41 (.19) & .45 (.19) \\
	\cline{2-9}
	\multirow{3}{*}{\rotatebox{90}{$d=25$}} & \multirow{3}{*}{\rotatebox{90}{Rec.}}
			& & & & & .07 (.07) & .48 (.24) & .60 (.24) \\
		& 	& .45 (.32) & .82 (.14) & \!\!{\bf .84 (.12)}\!\! & \!\!{\bf .86 (.11)}\!\! 
					& .23 (.18) & .74 (.19) & .74 (.19) \\
		& 	& & & & & .80 (.20) & .83 (.13) & \!\!{\bf .86 (.11)}\!\! \\
	\cline{2-9}
		& \multirow{3}{*}{\rotatebox{90}{F}}
			& & & & & .09 (.08) & .41 (.21) & .55 (.23) \\
		& 	& .56 (.22) & \!\!{\bf .75 (.14)}\!\! & \!\!{\bf .66 (.17)}\!\! & \!\!{\bf .60 (.19)}\!\! 
					& .21 (.16) & .53 (.21) & .58 (.20) \\
		& 	& & & & & .45 (.18) & .53 (.19) & .58 (.18) \\
	\cline{2-9}
	& \rotatebox{90}{${\rm F}_0$}
		& .92 (.02) & .92 (.02) & .92 (.02) & .92 (.02) & .92 (.02) & .93 (.02) & .92 (.02) \\
	\hline
	\hline
		& \multirow{3}{*}{\rotatebox{90}{Prec.}}
			& & & & & .10 (.13) & .24 (.20) & \!\!{\bf .58 (.19)}\!\! \\
		& 	& \!\!{\bf .87 (.11)}\!\! & \!\!{\bf .69 (.14)}\!\! & .56 (.17) & .47 (.17) 
					& .13 (.14) & .37 (.20) & .52 (.19) \\
		& 	& & & & & .27 (.19) & .42 (.18) & .47 (.18) \\
	\cline{2-9}
	\multirow{3}{*}{\rotatebox{90}{$d=50$}} & \multirow{3}{*}{\rotatebox{90}{Rec.}}
			& & & & & .04 (.04) & .18 (.19) & .60 (.19) \\
		& 	& .41 (.20) & .83 (.11) & \!\!{\bf .85 (.10)}\!\! & \!\!{\bf .91 (.05)}\!\! 
					& .10 (.11) & .51 (.21) & .72 (.16) \\
		& 	& & & & & .50 (.22) & .81 (.12) & \!\!{\bf .86 (.08)}\!\! \\
	\cline{2-9}
		& \multirow{3}{*}{\rotatebox{90}{F}}
			& & & & & .05 (.06) & .20 (.19) & .58 (.19) \\
		& 	& .53 (.20) & \!\!{\bf .75 (.12)}\!\! & \!\!{\bf .66 (.15)}\!\! & \!\!{\bf .61 (.15)}\!\! 
					& .10 (.11) & .42 (.20) & .59 (.18) \\
		& 	& & & & & .34 (.20) & .54 (.17) & .60 (.16) \\
	\cline{2-9}
	& \rotatebox{90}{${\rm F}_0$}
		& .90 (.03) & .90 (.02) & .89 (.02) & .89 (.03) & .89 (.03) & .90 (.02) & .90 (.03) \\
	\hline
	\hline
		& \multirow{3}{*}{\rotatebox{90}{Prec.}}
			& & & & & .09 (.11) & .17 (.14) & \!\!{\bf .68 (.15)}\!\! \\
		& 	& \!\!{\bf .91 (.07)}\!\! & \!\!{\bf .78 (.10)}\!\! & .64 (.14) & .53 (.15) 
					& .10 (.12) & .33 (.21) & .62 (.16) \\
		& 	& & & & & .22 (.17) & .46 (.18) & .55 (.16) \\
	\cline{2-9}
	\multirow{3}{*}{\rotatebox{90}{$d=100$}} & \multirow{3}{*}{\rotatebox{90}{Rec.}}
			& & & & & .03 (.10) & .06 (.10) & .59 (.17) \\
		& 	& .37 (.18) & .81 (.11) & \!\!{\bf .83 (.11)}\!\! & \!\!{\bf .95 (.02)}\!\! 
					& .06 (.10) & .25 (.21) & .67 (.15) \\
		& 	& & & & & .24 (.19) & .67 (.16) & \!\!{\bf .82 (.09)}\!\! \\
	\cline{2-9}
		& \multirow{3}{*}{\rotatebox{90}{F}}
			& & & & & .05 (.10) & .08 (.11) & .63 (.16) \\
		& 	& .51 (.19) & \!\!{\bf .79 (.10)}\!\! & \!\!{\bf .72 (.12)}\!\! & \!\!{\bf .67 (.12)}\!\! 
					& .07 (.10) & .28 (.21) & .64 (.15) \\
		& 	& & & & & .22 (.18) & .54 (.17) & .65 (.14) \\
	\cline{2-9}
	& \rotatebox{90}{${\rm F}_0$}
		& .87 (.04) & .87 (.04) & .87 (.03) & .87 (.03) & .87 (.03) & .88 (.04) & .87 (.03) \\
	\hline
\end{tabular}
\end{table*}

We conducted simulations for three cases with data dimensionality $d = 25, 50$ and $100$ where the number of datasets is fixed at $N = 5$.
For each case, we generate precision matrices $\Lm_1, \Lm_2, \ldots, \Lm_N$ to have $15\%$ non-zero entries on average.
In the simulation, we randomly generate datasets $100$ times and applied each method using several different hyper-parameters, 
where in each run we set the number of data points in each dataset to be $5d$.
For CSSL, we use the heuristic with a parameter $\alpha$ varying from $10^{-2}$ to $10^{-0}$ over $41$ values.
We also evaluate results for $\rho = \alpha$ and $\gm = \infty$ to see the effect of $\gm$ in an extreme case.
As discussed in Section \ref{sec:interpret_cssl}, this corresponds to solving a single SICS problem with $S = \sum_{i=1}^N t_i S_i$
and setting the result to $\hat{\Lm}_1 = \hat{\Lm}_2 = \ldots = \hat{\Lm}_N = \hat{\Lm}$.
For SICS and MSICS, we set the value of $\rho$ as $\rho = \alpha$.
For each method, we adopt the resulting precision matrices with $15\%$ non-zero entries among these $41$ values of $\alpha$.
In SICS and MSICS, we also vary the thresholding parameter $\epsilon_0$ between $0.5, 0.7$ and $0.9$.

We summarize the results in \tablename~\ref{tab:simu}.
From the table, we can see the clear advantage of CSSL for $p=2$ and $\infty$ over the other methods.
These two methods show higher F-measures, which are from their higher precision and recall.
This contrasts with other methods, SICS and MSICS, which achieve high recall, but have relatively poor precision.
This means that structure detected by those methods involve not only true common substructure but also many false detections.
This shows the drawback of estimated precision matrices derived through SICS and MSICS, that is, 
their estimators tend to be highly varied even for true common entries while this is not the case for CSSL.
This phenomenon is especially significant in SICS, which can hardly find common substructures owing to its highly varied estimators.
The results for MSICS under $p = \infty$ and $\epsilon_0 = 0.9$ are still better than the others, 
although $\epsilon_0 = 0.9$ means that $90\%$ of estimated non-zero entries are considered common, which is too optimistic.
Moreover, we can see that the improvement of the F-measure is achieved by the growth of recall
by contrasting the results with $\epsilon_0 = 0.5$ and $0.9$.
This means that variations on the true common substructure mostly happens in between $50\%$ and $90\%$ 
of the entire variations of the estimated precision matrices, which are highly varied and can hardly be considered common.
Note that despite the significant difference in the common entry detection performance, 
all methods achieve comparable zero pattern identification performance as shown by the ${\rm F}_0$-measure.
This shows that finding common entries is a different problem from the ordinal graphical model selection, and that only CSSL does well at both tasks.

We note that CSSL with $p = 1$ and $\gm = \infty$ give two extreme results.
In the former setting, the resulting precision matrices achieve higher precision with lower recall, 
which is very conservative, while it is the opposite in the latter setting.
The first result is caused by the difference of a grouped regularization $\pnorm{\Om}{1, p}$ for $p = 1$ and $p > 1$.
For $p = 1$, $\pnorm{\Om}{1, p}$ completely decouples into ordinary $\ell_1$-regularizations 
and the resulting precision matrices do not necessarily have common zero entries in individual substructures.
Intuitively speaking, the results for $p = 1$ have common zero entries
$\Om_{1, jj'} = \Om_{2, jj'} = \ldots = \Om_{N, jj'} = 0$ only when it is strongly confident, 
which results in a very conservative performance compared with $p > 1$.
On the other hand, if $\gm = \infty$, the entire structures are considered to be common, which results in fewer false negatives and more false positives.

%==============================================================================================
\section{Application to Anomaly Detection}
\label{sec:anom}

In this section, we apply CSSL to an anomaly detection problem.
The task is to identify contributions of each variable to the difference between two datasets.
Correlation anomalies~\citep{ide2009proximity}, or errors on dependencies between variables, 
are known to be difficult to detect using existing approaches, especially with noisy data.
To overcome this problem, the use of sparse precision matrices was proposed by~\cite{ide2009proximity}, 
since the sparse approach reasonably suppresses the pseudo-correlation among variables
caused by noise and improves the detection rate.
Here, we propose using CSSL.
There is a clear indication that the proposed method can further suppress the variation in the estimated matrices.
In particular, we expect that dependency structures among healthy variables are estimated to be common,
which reduces the risk that such variables are mis-detected and only anomalies are enhanced.

%==============================================================================================
\subsection{Anomaly Score}

We adopt the measurement for correlation anomalies proposed by~\cite{ide2009proximity}.
This score is based on the KL-divergence between two conditional distributions.
Formally, let $\Vec{x}^{\rm A}, \Vec{x}^{\rm B} \in \R^d$ be Gaussian random variables
following $\ND(\Vec{0}_d, {\Lm^{\rm A}}^{-1})$ and $\ND(\Vec{0}_d, {\Lm^{\rm B}}^{-1})$, respectively.
We measure the degree of anomaly between their $j$th variables $x_j^{\rm A}$ and $x_j^{\rm B}$
using a KL-divergence between their conditional distributions 
$p_{\rm A}(x_j^{\rm A} | \Vec{x}_{\backslash j}^{\rm A})$ and $p_{\rm B}(x_j^{\rm B} | \Vec{x}_{\backslash j}^{\rm B})$, 
where $\Vec{x}_{\backslash j}^{\rm A}$ and $\Vec{x}_{\backslash j}^{\rm B}$ are the remaining $d-1$ variables.
To compute the score, we first divide the precision matrix $\Lm^{\rm A}$ and its inverse $W^{\rm A}$ into a
$(d-1) \times (d-1)$ dimensional matrix, a $d-1$ dimensional vector, and a scalar, 
\begin{align*}
	\Lm^{\rm A} = \matrix{cc}{L^{\rm A}_{\backslash j} & \Vec{l}^{\rm A}_{\backslash j} \\ \Vec{l}^{\rm A}_{\backslash j} & \lambda^{\rm A}_j}
		\; , \;\; W^{\rm A} = {\Lm^{\rm A}}^{-1}= \matrix{cc}{V^{\rm A}_{\backslash j} & \Vec{v}^{\rm A}_{\backslash j} \\
		\Vec{v}^{\rm A}_{\backslash j} & \sigma^{\rm A}_j} \, ,
\end{align*}
where we have rotated the rows and columns of $\Lm^{\rm A}$ and $W^{\rm A}$ simultaneously so that their original $j$th rows and columns
are located at the last rows and columns of the matrix.
The matrices $\Lm^{\rm B}$ and its inverse $W^{\rm B}$ are also divided in a same manner.
The score is then given as
\begin{align*}
	d_j^{\rm AB} & = \int d\Vec{x}_{\backslash j}^{\rm A} \, p_{\rm A}(\Vec{x}_{\backslash j}^{\rm A}) \, 
		\KL(p_{\rm A}(x_j^{\rm A} | \Vec{x}_{\backslash j}^{\rm A}) || p_{\rm B}(x_j^{\rm B} | \Vec{x}_{\backslash j}^{\rm B})) \\
	& = {\Vec{v}^{\rm A}_{\backslash j}}^\top (\Vec{l}^{\rm A}_{\backslash j} - \Vec{l}^{\rm B}_{\backslash j})
		+ \frac{1}{2} \left\{\frac{{\Vec{l}^{\rm B}_{\backslash j}}^\top V^{\rm B}_{\backslash j} \Vec{l}^{\rm B}_{\backslash j}}
		{\lambda^{\rm B}_j} - \frac{{\Vec{l}^{\rm A}_{\backslash j}}^\top V^{\rm A}_{\backslash j} \Vec{l}^{\rm A}_{\backslash j}}
		{\lambda^{\rm A}_j} \right\} \\
	& \hspace{12pt}	+ \frac{1}{2} \left\{ \ln \frac{\lambda^{\rm A}_j}{\lambda^{\rm B}_j}
		+ \sigma^{\rm A}_j(\lambda^{\rm A}_j - \lambda^{\rm B}_j) \right\} \, .
\end{align*}
Here, the KL-divergence is averaged over the remaining $d-1$ variables $\Vec{x}^{\rm A}_{\backslash j}$.
Since the KL-divergence is not symmetric and $d_j^{\rm AB} \neq d_j^{\rm BA}$ holds in general, 
the resulting anomaly score $a_j$ is decided as their maximum: 
\begin{align*}
	a_j = \max (d_j^{\rm AB}, d_j^{\rm BA}) \, .
\end{align*}

%==============================================================================================
\subsection{Simulation Setting}

We evaluate the anomaly detection performance using {\it sensor error} data~\citep{ide2009proximity}.
The dataset comprised 42 sensor values collected from a real car in 79 normal states and 20 faulty states.
The fault is caused by mis-wiring of the 24th and 25th sensors, resulting in correlation anomalies.
Since sample covariances are rank-deficient in some datasets, we added $10^{-3}$ on their diagonal to avoid singularities.

For simulation, we randomly sample $n_{\rm n}$ datasets from the normal states and $n_{\rm f}$ datasets from the faulty states,
and then estimate sparse precision matrices using six methods, CSSL with $p = 1, 2$ and $\infty$, SICS~(\ref{eq:sics}), 
and MSICS~(\ref{eq:msics}) with $p = 2$ and $\infty$.
For CSSL, we adopt the heuristic and set $\rho = \max (s_1 \alpha + s_0, 0)$ and $\gm = \alpha$ for a given $\alpha$, 
and for SICS and MSICS, we set $\rho = \alpha$.
We test each method for $11$ different values of $\alpha$ ranging from $10^{-1.5}$ to $10^{-0.5}$.
The weight parameters $t_i$ in CSSL and MSICS are set as $t_i = \frac{1}{2 n_{\rm n}}$ for normal datasets
and $t_i = \frac{1}{2 n_{\rm f}}$ for faulty datasets to balance the effects from the two states.
Since the anomaly score is designed only for a pair of datasets, we calculate anomaly scores for each of $n_{\rm n} \times n_{\rm f}$ pairs of datasets.

%==============================================================================================
\subsection{Result}

\begin{table*}[t]
\centering
\caption{Anomaly detection results:
The simulation is conducted for 4 different settings, $[n_{\rm n}, n_{\rm f}] = [4, 1], [12, 3], [20, 5]$ and $[40, 10]$.
For each method, we compute precision matrices for $11$ different values of $\alpha$ ranging from $10^{-1.5}$ to $10^{-0.5}$.
The table shows the median of the best AUCs among these $11$ results over 100 random realizations of datasets.
The numbers in brackets are $25\%$ and $75\%$ quantiles.
The bold font represents the top three results, which are CSSL ($p = 2$), CSSL ($p = \infty$) and MSICS ($p = \infty$) for all settings.}
\label{tab:anom}
\begin{tabular}{|c||c|c||c|c|}
	\hline
	& \multicolumn{2}{|c||}{$[n_{\rm n}, n_{\rm f}] = [4, 1]$} 
	& \multicolumn{2}{|c|}{$[n_{\rm n}, n_{\rm f}] = [12, 3]$} \\
	\hline
	 & best AUC & $\alpha$ & best AUC & $\alpha$ \\
	\hline
	\hspace{-7pt} CSSL ($p = 1$) \hspace{-7pt} 
		& .975 (.950 / .987)		& \hspace{-6pt} $10^{-0.9}$ \hspace{-6pt} 
		& .975 (.950 / 1.00)		& \hspace{-6pt} $10^{-0.9}$ \\
	\hspace{-7pt} CSSL ($p = 2$) \hspace{-7pt}
		& {\bf .987 (.963 / 1.00)}	& \hspace{-6pt} $10^{-0.9}$ \hspace{-6pt} 
		& {\bf .987 (.963 / 1.00)}	& \hspace{-6pt} $10^{-0.9}$ \\
	\hspace{-7pt} CSSL ($p = \infty$) \hspace{-7pt}
		& {\bf .987 (.963 / 1.00)}	& \hspace{-6pt} $10^{-0.9}$ \hspace{-6pt} 
		& {\bf 1.00 (.987 / 1.00)}	& \hspace{-6pt} $10^{-0.9}$ \\
	SICS
		& .975 (.938 / .987)		& \hspace{-6pt} $10^{-0.5}$ \hspace{-6pt} 
		& .975 (.938 / .987)		& \hspace{-6pt} $10^{-0.5}$ \\
	\hspace{-7pt} MSICS ($p = 2$) \hspace{-7pt}
		& .975 (.950 / .987)		& \hspace{-6pt} $10^{-0.8}$ \hspace{-6pt} 
		& .975 (.950 / .987)		& \hspace{-6pt} $10^{-0.7}$ \\
	\hspace{-7pt} MSICS ($p = \infty$) \hspace{-7pt}
		& {\bf .987 (.963 / 1.00)}	& \hspace{-6pt} $10^{-1.1}$ \hspace{-6pt} 
		& {\bf .987 (.975 / 1.00)}	& \hspace{-6pt} $10^{-1.2}$ \\
	\hline
\end{tabular} \\
\hspace{1pt}
\begin{tabular}{|c||c|c||c|c|}
	\hline
	& \multicolumn{2}{|c||}{$[n_{\rm n}, n_{\rm f}] = [20, 5]$} 
	& \multicolumn{2}{|c|}{$[n_{\rm n}, n_{\rm f}] = [40, 10]$} \\
	\hline
	 & best AUC & $\alpha$ & best AUC & $\alpha$ \\
	\hline
	\hspace{-7pt} CSSL ($p = 1$) \hspace{-7pt} 
		& .975 (.950 / 1.00)		& \hspace{-6pt} $10^{-0.9}$ \hspace{-6pt} 
		& .975 (.963 / 1.00)		& \hspace{-6pt} $10^{-0.9}$ \\
	\hspace{-7pt} CSSL ($p = 2$) \hspace{-7pt}
		& {\bf 1.00 (.975 / 1.00)}	& \hspace{-6pt} $10^{-0.8}$ \hspace{-6pt} 
		& {\bf .987 (.963 / 1.00)}	& \hspace{-6pt} $10^{-0.8}$ \\
	\hspace{-7pt} CSSL ($p = \infty$) \hspace{-7pt}
		& {\bf 1.00 (.987 / 1.00)}	& \hspace{-6pt} $10^{-0.9}$ \hspace{-6pt} 
		& {\bf 1.00 (.987 / 1.00)}	& \hspace{-6pt} $10^{-0.9}$ \\
	SICS
		& .975 (.950 / .987)		& \hspace{-6pt} $10^{-0.5}$ \hspace{-6pt} 
		& .975 (.950 / .987)		& \hspace{-6pt} $10^{-0.5}$ \\
	\hspace{-7pt} MSICS ($p = 2$) \hspace{-7pt}
		& .975 (.950 / .987)		& \hspace{-6pt} $10^{-1.0}$ \hspace{-6pt} 
		& .975 (.950 / .987)		& \hspace{-6pt} $10^{-1.0}$ \\
	\hspace{-7pt} MSICS ($p = \infty$) \hspace{-7pt}
		& {\bf .987 (.975 / 1.00)}	& \hspace{-6pt} $10^{-1.1}$ \hspace{-6pt} 
		& {\bf .987 (.975 / 1.00)}	& \hspace{-6pt} $10^{-0.9}$ \\
	\hline
\end{tabular}
\vspace{-2.5ex}
\end{table*}

\begin{figure*}[t]
\centering
\subfigure[CSSL ($p = 1$)]{
\includegraphics[width=0.27\textwidth]{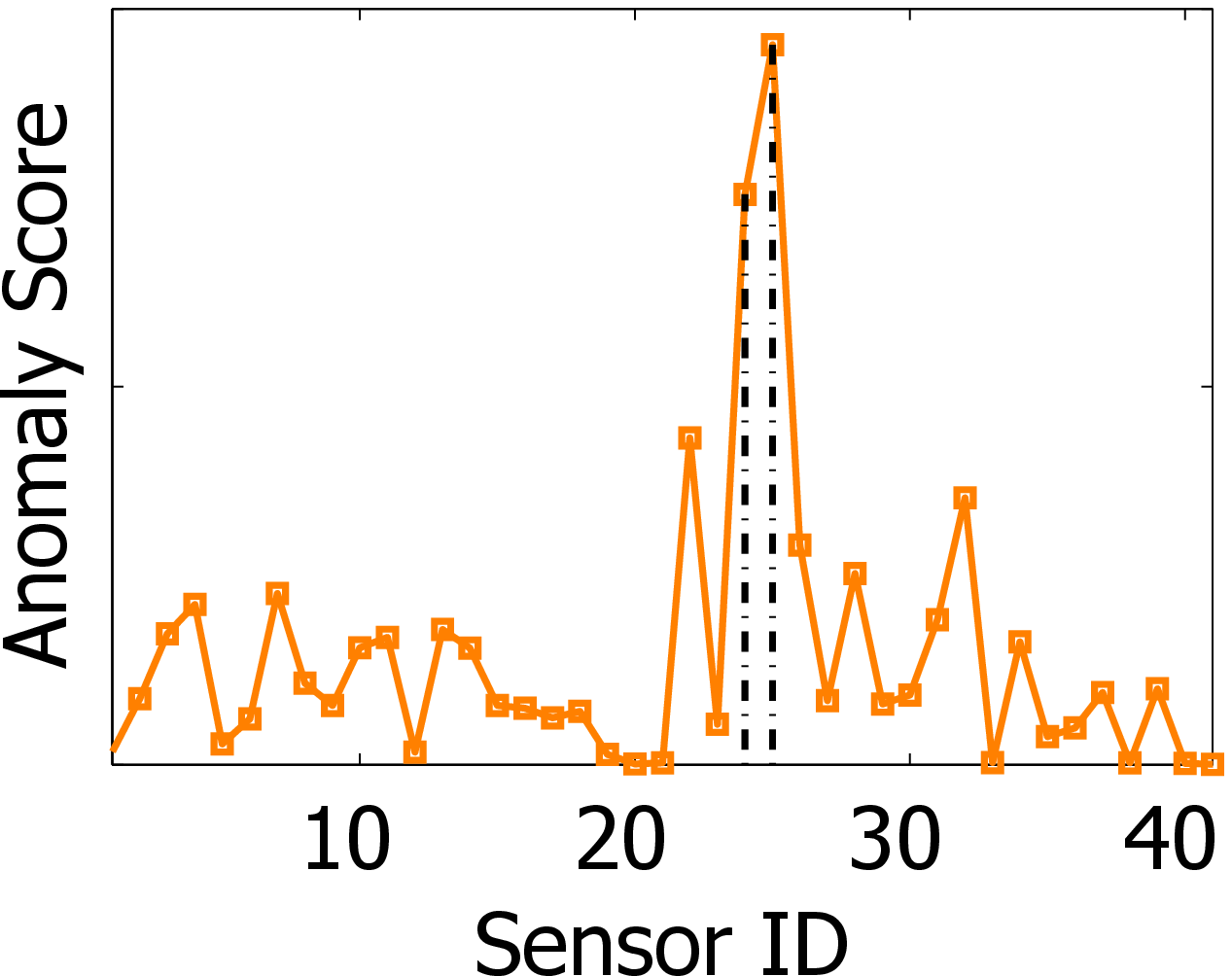}
\label{fig:anom_cssl_1}}
\hspace{5pt}
\subfigure[CSSL ($p = 2$)]{
\includegraphics[width=0.27\textwidth]{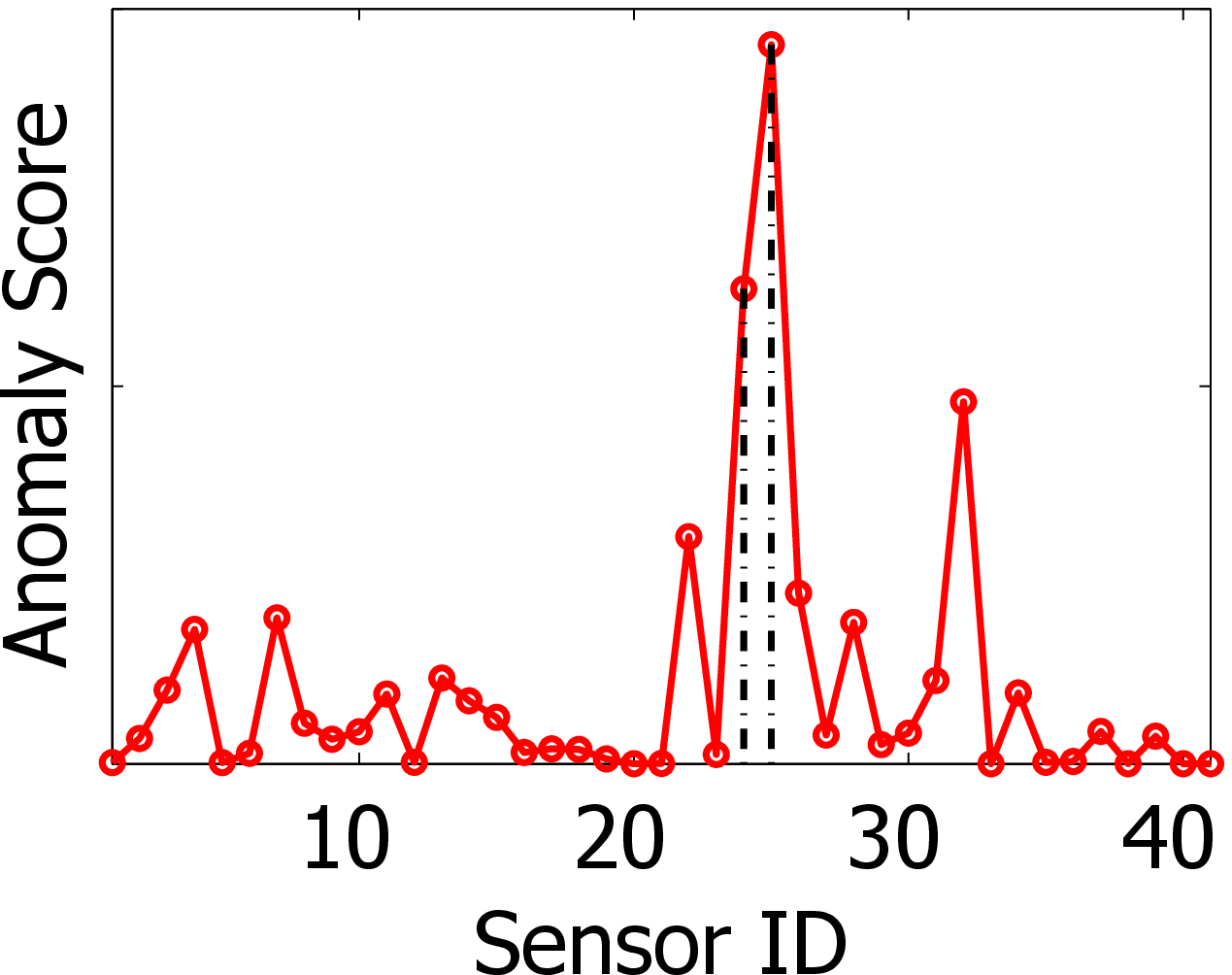}
\label{fig:anom_cssl_2}}
\hspace{5pt}
\subfigure[CSSL ($p = \infty$)]{
\includegraphics[width=0.27\textwidth]{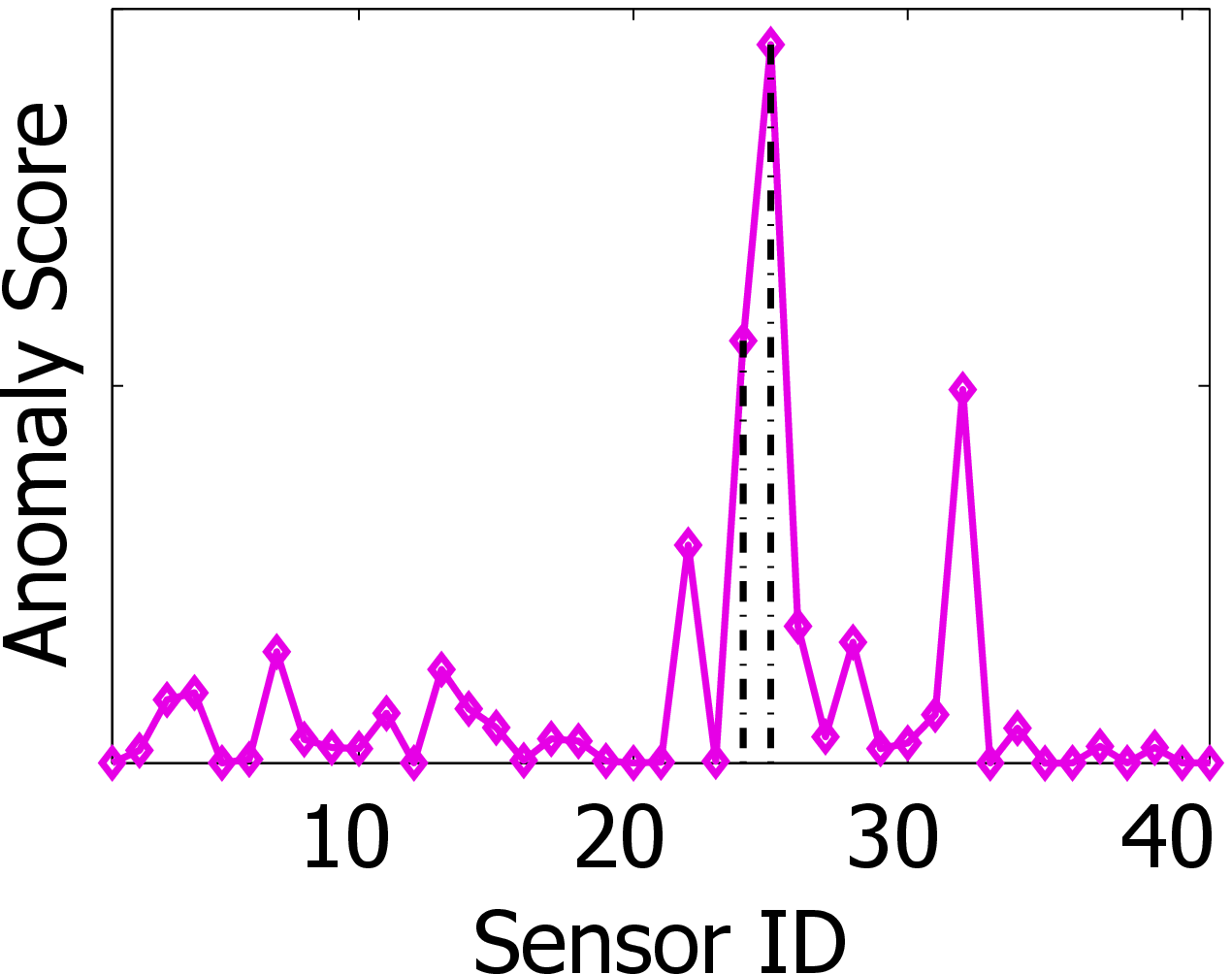}
\label{fig:anom_cssl_inf}} \\
\subfigure[SICS]{
\includegraphics[width=0.27\textwidth]{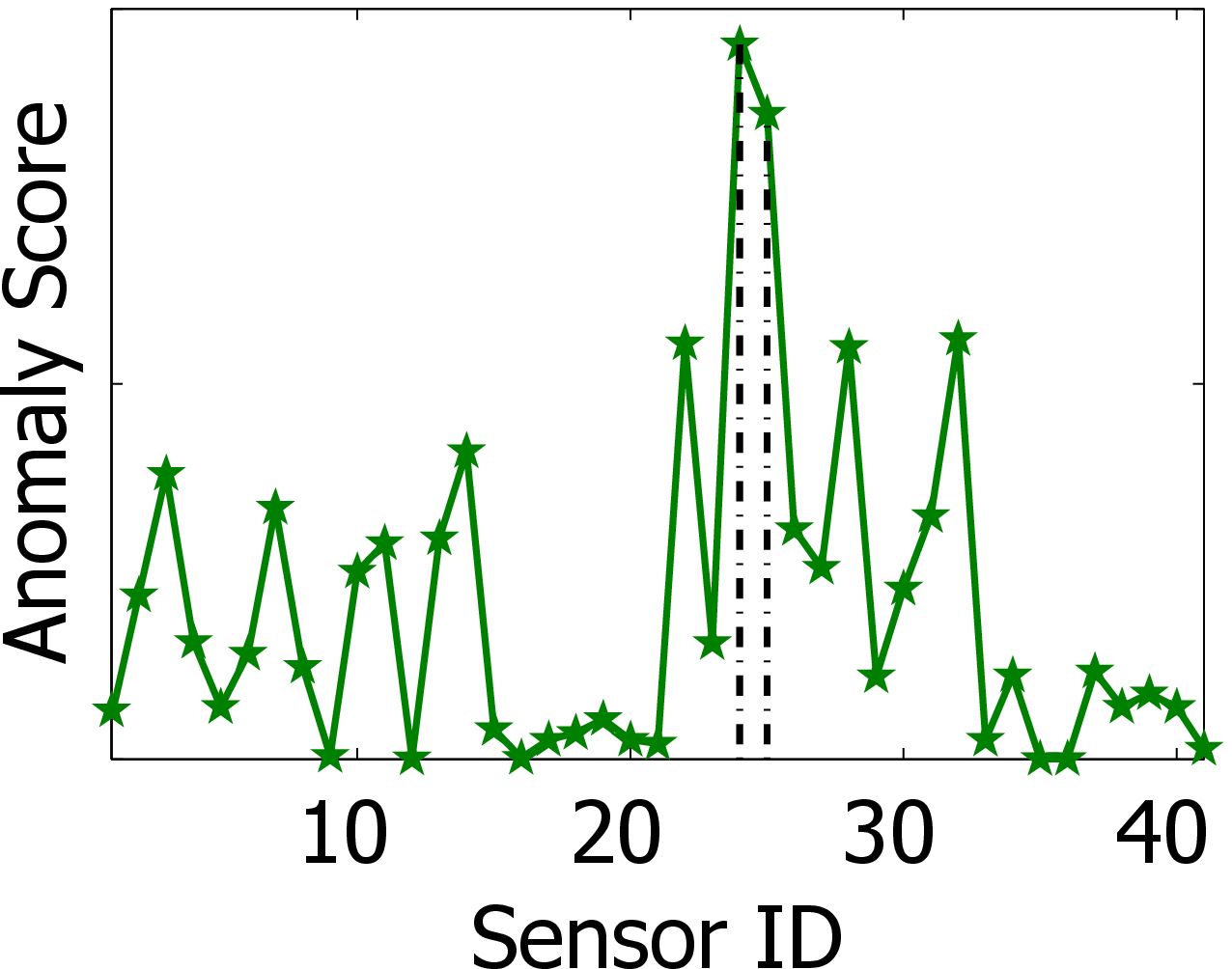}
\label{fig:anom_sics}}
\hspace{5pt}
\subfigure[MSICS ($p = 2$)]{
\includegraphics[width=0.27\textwidth]{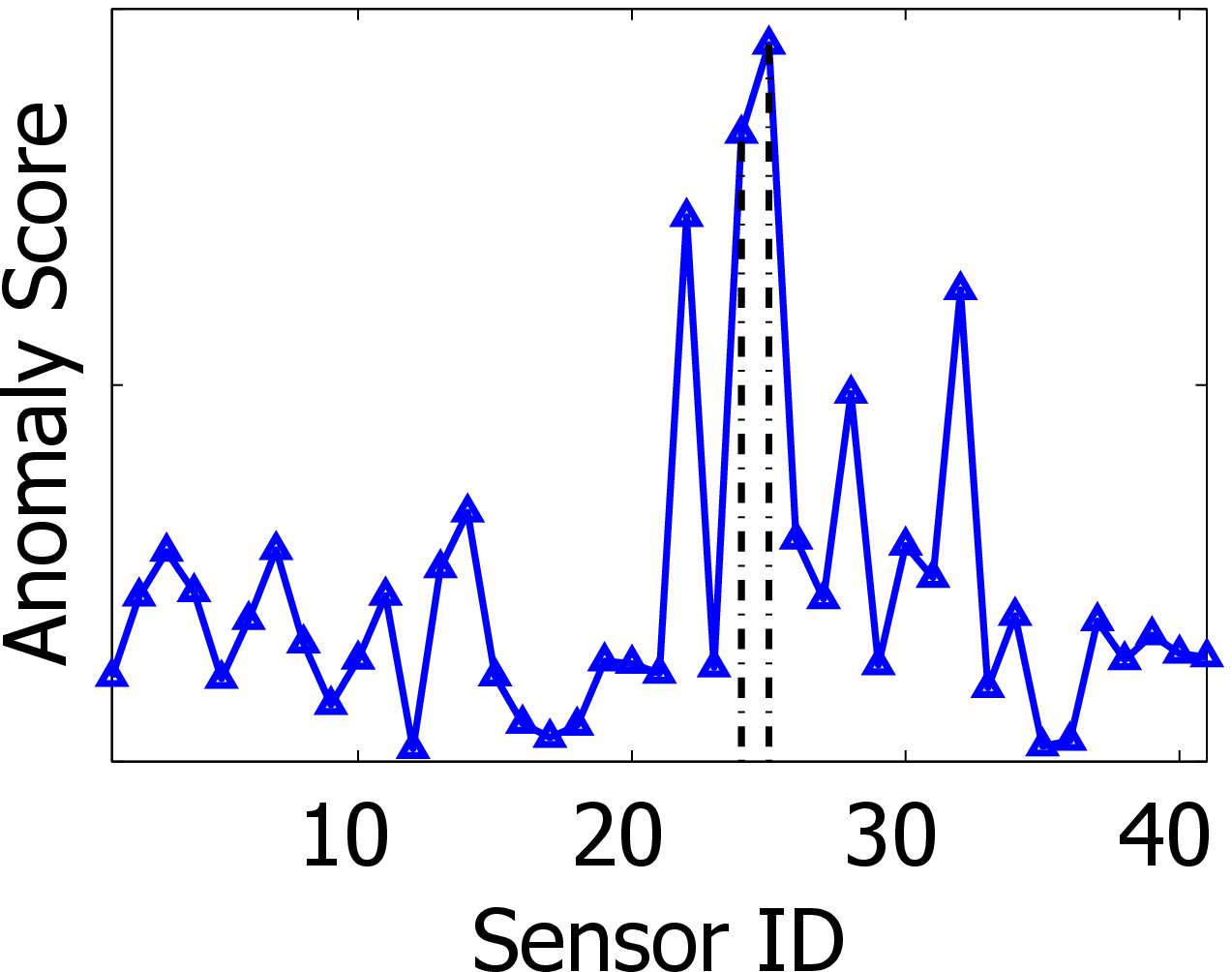}
\label{fig:anom_msics_2}}
\hspace{5pt}
\subfigure[MSICS ($p = \infty$)]{
\includegraphics[width=0.27\textwidth]{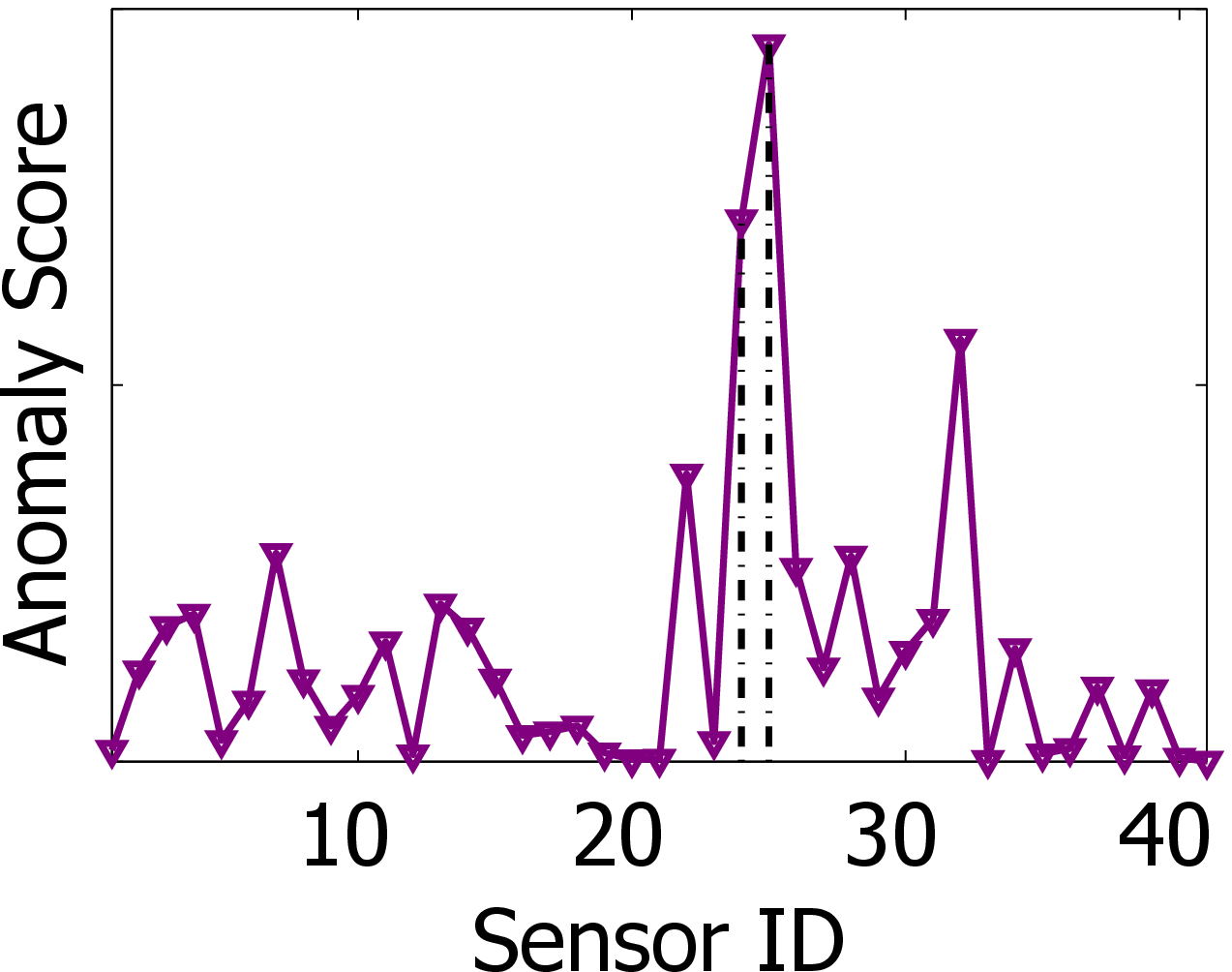}
\label{fig:anom_msics_inf}}
\caption{Median anomaly scores for each method for $[n_{\rm n}, n_{\rm f}] = [20, 5]$ with best AUCs. 
Each plot is normalized so that the maximum is the same.
Dotted lines denote true faulty sensors.}
\label{fig:anom}
\vspace{-3ex}
\end{figure*}

We repeated the above procedure 100 times for 4 different settings, $[n_{\rm n}, n_{\rm f}]$ $= [4, 1], [12, 3], [20, 5]$ and $[40, 10]$.
For each run, we evaluated the detection performance of each method by drawing an ROC curve and measuring the area under the curve (AUC).
In \tablename~\ref{tab:anom}, we summarize the best median results for each method and setting.
The table shows that CSSL with $p = 2, \infty$ and MSICS with $p = \infty$ achieve better detection performances than the others.
In particular, CSSL with $p = 2$ and $\infty$ achieve AUC = 1 as their median performance in some cases.
This means that they detect faulty sensors perfectly for more than half of the simulation.
To see further differences, we plot the median anomaly scores derived from each method 
for $[n_{\rm n}, n_{\rm f}] = [20, 5]$ in \figurename~\ref{fig:anom}.
From these graphs, we observe a clear distinction between successful methods and other methods on the significance of healthy sensors.
The 22nd and 28th sensors are relatively highly enhanced in SICS and MSICS with $p = 2$, but are not in CSSL and MSICS with $p = \infty$.
We conjecture that this is the major cause of performance differences.
Interestingly, not only the 22nd and 28th sensors but most of the other healthy sensors also have the same tendencies.
That is, CSSL and MSICS with $p = \infty$ reasonably suppress their significance while keeping erroneous sensors enhanced.
Moreover, although the differences are subtle, we can see that CSSL with $p = 2$ and $\infty$ more successfully
suppress the significance of sensors 1 to 21 and 33 to 42 than does MSICS with $p = \infty$.
Thus, as we expected in the beginning, CSSL reduces the nuisance effects and highlights only those variables with correlation anomalies.
The remaining peaks at some healthy variables are caused by the effect of the two faulty sensors 
since their effects may propagate to other healthy yet highly related sensors.

%==============================================================================================
\section{Conclusion}
\label{sec:concl}

In this paper, we formulated the CSSL problem for multiple GGMs.
We further provided a simple DAL-ADMM algorithm where each update step can be solved in a very efficient manner.
Numerical results on synthetic datasets indicate the clear advantage of the CSSL approach, 
in that it can achieve high precision and recall at the same time, which existing GGM structure learning methods can not achieve.
We also applied the proposed CSSL technique to the anomaly detection task in {\it sensor error} data.
Through the simulation, we observed that CSSL could efficiently suppress nuisance effects among variables in noisy sensors
and successfully enhanced target faulty sensors.

Several future research topics have been indicated, including analyzing the asymptotic property of the CSSL problem~(\ref{eq:cssl}), 
and extending the current formulation to the adaptive Lasso~\citep{zou2006adaptive,fan2009network} type one to guarantee
the {\it oracle property}~\citep{zou2006adaptive} of the estimator.
Applying the notion of commonness to more general dependency models, 
such as those with non-linear relations or commonness based on higher-order moment statistics, is also important.

%==============================================================================================
\section*{Acknowledgments}

We would like to acknowledge support for this project from the
JSPS Grant-in-Aid for Scientific Research(B) \#22300054.
The authors would like to thank Tsuyoshi Id\'e and his colleagues for providing {\it sensor error} datasets for our simulation.
We also received several helpful comments from Shohei Shimizu.


\appendix

%==============================================================================================
\section{Solutions to~(\ref{eq:yupdate2}) for $q = 1, 2$ and $\infty$}
\label{app:yupdate}

Here, we give detailed derivations of \tablename~\ref{tab:yupdate}.

%==============================================================================================
\subsection{The solution is in $\pC_1$.}
\label{app:c1}

Problem~(\ref{eq:yupdate_sub}) for $\Vec{y} \in \pC_1$ is formulated as follows: 
\begin{align}
	\min_{\Vec{y}} \frac{1}{2} \pnorm{\Vec{y} - \Vec{y}_0}{2}^2 \st |\Vec{1}_N^\top \Vec{y}| = \rho \, .
	\label{eq:yupdate_sub1}
\end{align}
Note that we have ignored the constraint $\pnorm{\Vec{y}}{q} \neq \gm$ because it holds for general $\Vec{y}_0$ and $\gm$ with probability one.
Hence, our interest is whether the solution to~(\ref{eq:yupdate_sub1}) satisfies $\pnorm{\Vec{y}}{q} \leq \gm$ or not. 
The additional constraint is not important in this respect.

The problem~(\ref{eq:yupdate_sub1}) has two possible cases as its solution, $\Vec{1}_N^\top \Vec{y} = \rho$ and $\Vec{1}_N^\top \Vec{y} = -\rho$.
For each case, we can solve the problem using Lagrange multipliers: 
\begin{align*}
	\min_{\Vec{y}} \max_\mu \frac{1}{2} \pnorm{\Vec{y} - \Vec{y}_0}{2}^2 + \mu (\Vec{1}_N^\top \Vec{y} - \zeta) \, ,
\end{align*}
where $\zeta \in \{\rho, -\rho \}$.
By setting the derivative over $\Vec{y}$ to zero, we get $\Vec{y} = \Vec{y}_0 - \mu \Vec{1}_N$. 
Moreover, by substituting this result above, we derive the optimal $\mu$ as $\mu = \frac{1}{N} (\Vec{1}_N^\top \Vec{y}_0 - \zeta)$
and the resulting objective function value is $\frac{1}{2N} \left( \Vec{1}_N^\top \Vec{y}_0 - \zeta \right)^2$.
The constraint $\zeta = \rho$ or $\zeta = -\rho$ is chosen so that this objective function value is minimized.
Obviously, $\zeta = \rho$ is optimal for the case when $\Vec{1}_N^\top \Vec{y}_0 \geq 0$, while $\zeta = -\rho$ for $\Vec{1}_N^\top \Vec{y}_0 < 0$.
Thus, the overall solution to problem~(\ref{eq:yupdate_sub1}) is
\begin{align*}
	\Vec{y} = \Vec{y}_0 - \frac{\Vec{1}_N^\top \Vec{y}_0 - \rho \, \sgn{\Vec{1}_N^\top \Vec{y}_0}}{N}\Vec{1}_N \, .
\end{align*}

%==============================================================================================
\subsection{The solution is in $\pC_2$ for $q = 1$.}
\label{app:c2_1}

When the solution is in $\pC_2$, the problem is formulated as
\begin{align}
	\min_{\Vec{y}} \frac{1}{2} \pnorm{\Vec{y} - \Vec{y}_0}{2}^2 \st \| \Vec{y} \|_q = \gm \, .
	\label{eq:yupdate_sub2}
\end{align}
Here, the shape of the constraint boundary changes according to the value of $q$.
For general $q \in [1, \infty]$, there exist several algorithms to solve this problem~\citep{boyd2004convex,sra2011fast}.
Especially, for $q = 1, 2$ and $\infty$, solutions are available in a very efficient manner.

For $q = 1$, it has been shown by \cite{honorio2010multi} that the problem is equivalent to the following {\it Continuous Quadratic Knapsack Problem}: 
\begin{align}
	& \min_{\Vec{z}} \sum_{i=1}^N \frac{1}{2} \left( z_i - |y_{0, i}| \right)^2 \st \Vec{z} \geq 0, \; \Vec{1}_N^\top \Vec{z} = \gm \, ,
	\label{eq:cqkp1}
\end{align}
which relates to $\Vec{y}$ by $y_i = \sgn{y_{0, i}}z_i$.
\cite{honorio2010multi} have also provided a solution technique for this problem.
From the KKT condition, the solution to (\ref{eq:cqkp1}) is $z_i(\nu) = \max (|y_{0, i}| - \nu, 0)$ for some constant $\nu$.
Moreover, the optimal $\nu$ satisfies $\Vec{1}_N^\top \Vec{z}(\nu) = \gm$.
Since $\Vec{1}_N^\top \Vec{z}(\nu)$ is a decreasing piecewise linear function with breakpoints $|y_{0, i}|$, 
we can find a minimum breakpoint $\nu_0$ that satisfies $\Vec{1}_N^\top \Vec{z}(\nu_0) \leq \gm$ by sorting the $N$ breakpoints.
The optimal $\nu$ is then given as
\begin{align*}
	\nu = \frac{\sum_{i \in \mathcal{I}_0} |y_{0, i}| - \gm }{ |\mathcal{I}_0| } \, ,
\end{align*}
where $\mathcal{I}_0 = \{ i ; |y_{0, i}| - \nu_0 \geq 0 \}$.
Note that the complexity of this algorithm is $\Order(N \log N)$ since we conduct a sorting of
$N$ values~\footnote{We can further reduce this to expected linear time complexity by introducing a randomized algorithm \citep{duchi2008efficient}.}.

%==============================================================================================
\subsection{The solution is in $\pC_2$ for $q = 2$.}
\label{app:c2_2}

The solution to problem~(\ref{eq:yupdate_sub2}) for $q=2$ is analytically available.
We solve the problem using Lagrange multipliers: 
\begin{align*}
	\min_{\Vec{y}} \max_\lambda \frac{1}{2} \pnorm{\Vec{y} - \Vec{y}_0}{2}^2 + \frac{\lambda}{2} (\pnorm{\Vec{y}}{2}^2 - \gm^2) \, .
\end{align*}
By setting the derivative over $\Vec{y}$ to zero, we get $\Vec{y} = \frac{1}{1 + \lambda} \Vec{y}_0$.
Moreover, from the constraint $\pnorm{\Vec{y}}{2} = \gm$, the solution is
\begin{align*}
	\Vec{y} = \frac{\gm}{\pnorm{\Vec{y}_0}{2}}\Vec{y}_0 \, .
\end{align*}

%==============================================================================================
\subsection{The solution is in $\pC_2$ for $q = \infty$.}
\label{app:c2_inf}

The solution of~(\ref{eq:yupdate_sub2}) for the case $q = \infty$ is much simpler.
The problem is just a box-constrained least squares, with solution
\begin{align*}
	y_i = \left\{\begin{array}{cl}
		\gm & \; (\text{if} \;\; y_{0, i} > \gm) \\
		y_{0, i} & \; (\text{if} \;\; |y_{0, i}| \leq \gm) \\
		-\gm & \; (\text{if} \;\; y_{0, i} < - \gm)
	\end{array}\right.
\end{align*}
which is equivalent to $y_i = \sgn{y_{0, i}} \min (|y_{0, i}|, \gm)$.

%==============================================================================================
\subsection{The solution is in $\pC_3$ for $q = 1$.}
\label{app:c3_1}

We provide the solution procedure for~(\ref{eq:yupdate_sub}) for $\Vec{y} \in \pC_3$ and $q = 1$ based on the next theorem.
\begin{theorem}
	\label{th:yupdate_c3}
	Let $\tilde{\Vec{y}}$ be the solution to problem~(\ref{eq:yupdate_sub}) for $\Vec{y} \in \pC_1$, 
	and suppose it is infeasible in the original problem~(\ref{eq:yupdate2}).
	Then, the solution to~(\ref{eq:yupdate_sub}) for $\Vec{y} \in \pC_3$ has same signs as $\tilde{\Vec{y}}$, 
	that is, $\tilde{y}_i y_i \geq 0$ for $1 \leq i \leq N$.
\end{theorem}

From this result, we can factorize the variable indices into two parts, $\mathcal{I}_+ = \{i; \tilde{y}_i \geq 0 \}$ 
and $\mathcal{I}_- = \{i; \tilde{y}_i < 0 \}$.
Using this factorization, the objective function is expressed as 
$\frac{1}{2} \sum_{i \in \mathcal{I}_+}(y_i-y_{0,i})^2+\frac{1}{2} \sum_{i \in \mathcal{I}_-}(y_i-y_{0,i})^2$.
The equality constraints can also be expressed as 
$\sum_{i \in \mathcal{I}_+} y_i + \sum_{i \in \mathcal{I}_-} y_i = \zeta$, with $\zeta \in \{\rho, -\rho \}$ and
$\sum_{i \in \mathcal{I}_+} y_i - \sum_{i \in \mathcal{I}_-} y_i = \gm$.
From these expressions, we derive two independent problems: 
\begin{align*}
	& \min_{\Vec{y}^+} \frac{1}{2} \sum_{i \in \mathcal{I}_+} \left( y_i^+ - y_{0,i} \right)^2
		\st \Vec{y}^+ \geq 0 \, , \; \sum_{i \in \mathcal{I}_+} y_i^+ = \frac{\gm + \zeta}{2} \, , \\
	& \min_{\Vec{y}^-} \frac{1}{2} \sum_{i \in \mathcal{I}_-} \left( y_i^- + y_{0,i} \right)^2
		\st \Vec{y}^- \geq 0 \, , \; \sum_{i \in \mathcal{I}_-} y_i^- = \frac{\gm - \zeta}{2} \, .
\end{align*}
The solutions to these problems relate to $\Vec{y}$ in that $y_i = y_i^+$ for $i \in \mathcal{I}_+$
and $y_i = - y_i^-$ for $i \in \mathcal{I}_-$.
These problems are continuous quadratic knapsack problems 
and the solution can be found by using the same algorithm as in problem~(\ref{eq:cqkp1}).
We derive the final solution by solving these problems for the two cases $\zeta = \rho$ and $\zeta = -\rho$, 
and choosing the one with the smaller objective function value in (\ref{eq:yupdate_sub}).

%==============================================================================================
\subsection{The solution is in $\pC_3$ for $q = 2$.}
\label{app:c3_2}

The solution in the case $\Vec{y} \in \pC_3$ and $q=2$ is analytically available. 
We use Lagrange multipliers: 
\begin{align*}
	\min_{\Vec{y}} \max_{\mu, \lambda} \frac{1}{2} \pnorm{\Vec{y} - \Vec{y}_0}{2}^2 
		+ \mu (\Vec{1}_N^\top \Vec{y} - \zeta) + \frac{\lambda}{2} (\pnorm{\Vec{y}}{2}^2 - \gm^2) \, ,
\end{align*}
where $\zeta \in \{ \rho, -\rho \}$.
By setting the derivative over $\Vec{y}$ to zero, we get $\Vec{y} = \frac{1}{1+\lambda}(\Vec{y}_0 - \mu \Vec{1}_N)$.
If $\rho = 0$, we have $\mu = \frac{\Vec{1}_N^\top \Vec{y}_0}{N}$ from the constraint $\Vec{1}_N^\top \Vec{y} = 0$.
Hence, from $\pnorm{\Vec{y}}{2} = \gm$, we get the optimal $\Vec{y}$ as
\begin{align*}
	\Vec{y} = \frac{\gm}{\pnorm{\Vec{y}_0 - \mu \Vec{1}_N}{2}} (\Vec{y} - \mu \Vec{1}_N) \, .
\end{align*}
For the case $\rho > 0$, we have $\frac{1}{1 + \lambda} = \frac{\zeta}{\Vec{1}_N^\top \Vec{y}_0 - N \mu}$ 
from the constraint $\Vec{1}_N^\top \Vec{y} = \zeta$.
Hence, we have a quadratic equation in $\mu$ from the constraint $\pnorm{\Vec{y}}{2}^2 = \gm^2$: 
\begin{align*}
	\rho^2 \pnorm{\Vec{v} - \mu \Vec{1}_N}{2}^2 = \gm^2 (\Vec{1}_N^\top \Vec{v} - N \mu)^2 \, .
\end{align*}
Solving this equation gives the optimal $\Vec{y}$ as
\begin{align*}
	\Vec{y} = \frac{\zeta}{\Vec{1}_N^\top \Vec{y}_0 - N \mu} (\Vec{y}_0 - \mu \Vec{1}_N) \, , \,\,
		\mu = \frac{1}{N} \left\{ \Vec{1}_N^\top \Vec{y}_0 \pm \sqrt{\tau} \right\} \, ,
\end{align*}
where $\tau = (\Vec{1}_N^\top \Vec{y}_0)^2 - N \frac{\gm^2 (\Vec{1}_N^\top \Vec{y}_0)^2 - \rho^2 \pnorm{\Vec{y}_0}{2}^2}{\gm^2 N - \rho^2}$.
By substituting this result into $\pnorm{\Vec{y} - \Vec{y}_0}{2}^2$, we have
\begin{align*}
	\hspace{-10pt}
	\pnorm{\Vec{y} - \Vec{y}_0}{2}^2 = \frac{1}{N} \left( \zeta - \Vec{1}_N^\top \Vec{y}_0 \right)^2 
		+ \frac{N \pnorm{\Vec{y}_0}{2}^2 - (\Vec{1}_N^\top \Vec{y}_0)^2}{N\tau} \left( \zeta \pm \sqrt{\tau} \right)^2 \, .
\end{align*}
Since $N \pnorm{\Vec{y}_0}{2}^2 - (\Vec{1}_N^\top \Vec{y}_0)^2 \geq 0$, the minimum of this value is achieved 
by choosing $\zeta$ and a sign in $\mu$ as $\zeta = \sgn{\Vec{1}_N^\top \Vec{y}_0}\rho$ and $-\sgn{\Vec{1}_N^\top \Vec{y}_0}$.
Thus, the overall result is
\begin{align*}
	& \Vec{y} = \sgn{\Vec{1}_N^\top \Vec{y}_0} \frac{\rho}{\Vec{1}_N^\top \Vec{y}_0 - N \mu} (\Vec{y}_0 - \mu \Vec{1}_N) \, , \\
	& \mu = \frac{1}{N} \left\{ \Vec{1}_N^\top \Vec{y}_0 
		- \sgn{\Vec{1}_N^\top \Vec{y}_0} \sqrt{ \tau } \right\} \, .
\end{align*}

%==============================================================================================
\subsection{The solution is in $\pC_3$ for $q = \infty$.}
\label{app:c3_inf}

The solution for~(\ref{eq:yupdate_sub}) with $\Vec{y} \in \pC_3$ and $q = \infty$ has two possible cases, 
$\Vec{1}_N^\top \Vec{y} = \rho$ and $\Vec{1}_N^\top \Vec{y} = - \rho$, where for each case the problem is:
\begin{align}
	\min_{\Vec{y}} \sum_{i=1}^N \frac{1}{2} \left( y_i - y_{0, i} \right)^2
		\st \Vec{1}_N^\top \Vec{y} = \zeta \; , \; - \gm \Vec{1}_N \leq \Vec{y} \leq \gm \Vec{1}_N \, ,
	\label{eq:cqkp2}
\end{align}
with $\zeta \in \{\rho, -\rho \}$.
Here, the constraint $\pnorm{\Vec{y}}{\infty} = \gm$ is relaxed to $\pnorm{\Vec{y}}{\infty} \leq \gm$.
However, if the solution to~(\ref{eq:cqkp2}) satisfies $\pnorm{\Vec{y}}{\infty} < \rho$, it has already been found as a solution to
(\ref{eq:yupdate_sub}) for $\Vec{y} \in \pC_1$ and therefore this relaxation does not affect the overall procedure.

Since problem (\ref{eq:cqkp2}) is a variant of the continuous quadratic knapsack problem, a similar strategy to (\ref{eq:cqkp1}) is available.
From the KKT condition, the solution to (\ref{eq:cqkp2}) is of the form 
$y_i(\nu) = \sgn{y_{0, i} - \nu} \min (|y_{0, i} - \nu|, \gm)$ for some constant $\nu$.
Moreover, the optimal $\nu$ satisfies $\Vec{1}_N^\top \Vec{y}(\nu) = \zeta$.
Since $\Vec{1}_N^\top \Vec{y}(\nu)$ is a decreasing piecewise linear function with breakpoints 
$\{ y_{0, i} - \gm, y_{0, i} + \gm \}_{i=1}^N$, 
we can find a minimum breakpoint $\nu_0$ that satisfies $\Vec{1}_N^\top \Vec{y}(\nu_0) \leq \zeta$ by sorting the $2N$ breakpoints.
The optimal $\nu$ is then
\begin{align*}
	\nu = \left\{\begin{array}{cl}
		\displaystyle{ \frac{ \sum_{i \in \mathcal{I}_2} y_{0, i} + \gm (|\mathcal{I}_1| - |\mathcal{I}_3|) - \zeta }{ |\mathcal{I}_2| } } 
			& (\text{if} \; \mathcal{I}_2 \neq \phi) \\
		\nu_0 & (\text{if} \; \mathcal{I}_2 = \phi)
	\end{array}\right.
\end{align*}
where $\mathcal{I}_1 = \{ i ;  y_{0, i} - \nu_0 \geq \gm \}$, $\mathcal{I}_2 = \{ i ; - \gm \leq y_{0, i} - \nu_0 < \gm \}$ 
and $\mathcal{I}_3 = \{ i ; y_{0, i} - \nu_0 < - \gm \}$.

%==============================================================================================
\section{Generation of Synthetic Precision Matrices}
\label{app:gen_data}

Here, we present the detailed procedure used to generate the sparse precision matrices with a common substructure in Section~\ref{sec:simu}.
The procedure is composed of two sequential steps. 
We first generate a single precision matrix, which is the common substructure in the resulting $N$ matrices.
Then, we add some non-zero entries to get a matrix $\Lm_i$.
This additional pattern is chosen to be unique for each matrix so that the resultant matrices $\Lm_1, \Lm_2, \ldots, \Lm_N$ satisfy 
the additive model assumption~(\ref{eq:prec_fact}).
In the following two subsections, we explain the above steps.

%==============================================================================================
\subsection{Generation of a Sparse Precision Matrix}

In several previous studies, synthetic sparse precision matrices are generated in a naive manner, that is, 
just adding a properly scaled identity matrix to a sparse symmetric matrix so that the resulting matrix 
is sparse and positive definite~\citep{banerjee2008model,wang2009solving,li2010inexact}.
In our simulations, we take a different approach to generating a sparse precision matrix for compatibility with the next step.

Our approach is based on an eigen-decomposition $\Lm = V D V^\top$, where $D$ is a matrix with eigenvalues on its diagonal
and $V$ is an orthonormal matrix such that $V^\top V = VV^\top = I_d$.
Here, we use the fact that $\Lm$ is sparse if $V$ is sufficiently sparse and the problem can be reduced to generating a sparse orthonormal matrix $V$.
This can be done easily by applying a Givens rotation~\citep{golub1996matrix} to an identity matrix $I_d$.
Formally, we let $V^{(0)} = I_d$ and apply the following procedure repeatedly until the desired sparsity is achieved.
\begin{enumerate}
	\item Randomly pick two indices $j, j'$ from $\{1, 2, \ldots, d\}$.
	\item Randomly generate $\theta$ from a uniform distribution $U([0, 2\pi])$.
	\item Update the $(j, j), (j, j'), (j', j)$ and $(j', j')$th entries of $V^{(k)}$ as
		\begin{align*}
			\hspace{-20pt}
			\matrix{cc}{V_{jj}^{(k+1)} & V_{jj'}^{(k+1)} \\ V_{j'j}^{(k+1)} & V_{j'j'}^{(k+1)}} \leftarrow
				\matrix{cc}{\cos \theta & -\sin \theta \\ \sin \theta & \cos \theta}
				\matrix{cc}{V_{jj}^{(k)} & V_{jj'}^{(k)} \\ V_{j'j}^{(k)} & V_{j'j'}^{(k)}} \, .
		\end{align*}
	\item Keep the remaining entries $V_{j_0 j'_0}^{(k+1)} \leftarrow V_{j_0 j'_0}^{(k)}$ 
		for $(j_0, j'_0)$ $\notin$ $\left\{ \right.$ $(j, j),$ $(j, j'),$ $(j', j),$ $(j', j')$ $\left. \right\}$.
\end{enumerate}
In our simulations, we generated each eigenvalue from a uniform distribution $U([0, 1])$.

%==============================================================================================
\subsection{Generation of Sparse Precision Matrices with a Common Substructure}

Here, we turn to imposing commonness on the resulting precision matrices.
To begin with, we generate small sparse precision matrices $\Psi_1, \Psi_2, \ldots, \Psi_a$ in the preceding manner
and construct a sparse block-diagonal precision matrix $\Lm_0 = \bdiag{\Psi_1, \Psi_2, \ldots, \Psi_a}$.
We then add some non-zero entries to $\Lm_0$ and generate $N$ precision matrices $\Lm_1, \Lm_2, \ldots, \Lm_N$.
At this stage, we keep the original non-zero entries $\Lm_0$ unchanged so they form a common substructure at the end.
Note that the addition of non-zero entries can not be done randomly since this might destroy the positive definiteness.

We describe the procedure for the case $a = 2$.
Let the eigen-decompositions of $\Psi_1$ and $\Psi_2$ be $\Psi_1 = V_1 D_1 V_1^\top$ and $\Psi_2 = V_2 D_2 V_2^\top$.
Note that $V_1$ and $V_2$ are sparse since they are generated to be so.
Now, let matrix $\Lm_i$ be of the form $\Lm_i = \matrix{cc}{\Psi_1 & \Phi_i \\ \Phi_i^\top & \Psi_2}$.
The objective is to generate a sparse non-zero matrix $\Phi_i$ while keeping the positive definiteness of $\Lm_i$.
This corresponds to keeping a determinant of $\Lm_i$ positive.
Here, we choose $\Phi_i$ of the form $\Phi_i = \tilde{V}_1^b \Xi_i \tilde{V}_2^{b \, \top}$
where $\Xi_i$ is a $b \times b$ diagonal matrix and $\tilde{V}_1^b$ and $\tilde{V}_2^b$ are matrices 
composed of $b$ columns in $V_1$ and $V_2$, respectively.
Specifically, we let $V_1 = \matrix{cccc}{\Vec{v}_{1, 1} & \Vec{v}_{1, 2} & \ldots & \Vec{v}_{1, d_1}}$ and
$V_2 = \matrix{cccc}{\Vec{v}_{2, 1} & \Vec{v}_{2, 2} & \ldots & \Vec{v}_{2, d_2}}$, 
where $d_1$ and $d_2$ denote the dimensionality of each matrix. 
Then $\tilde{V}_1^b$ and $\tilde{V}_2^b$ are 
$\tilde{V}_1^b = \matrix{cccc}{\Vec{v}_{1, \pi_{1, 1}} & \Vec{v}_{1, \pi_{1, 2}} & \ldots & \Vec{v}_{1, \pi_{1, b}}}$ and 
$\tilde{V}_2^b = \matrix{cccc}{\Vec{v}_{2, \pi_{2, 1}} & \Vec{v}_{2, \pi_{2, 2}} & \ldots & \Vec{v}_{2, \pi_{2, b}}}$, respectively, 
for some index sets $\{\pi_{1, 1}, \pi_{1, 2}, \ldots, \pi_{1, b} \} \subseteq \{1, 2, \ldots, d_1 \}$, 
$\left\{ \right.$ $\pi_{2, 1},$ $\left. \pi_{2, 2}, \ldots, \pi_{2, b} \} \subseteq \{1, 2, \ldots, d_2 \right\}$.
Then, from a general matrix property, we can express the determinant of $\Lm_i$ as
\begin{align*}
	\det \Lm_i & = \det \left(\Psi_1 - \Phi_i \Psi_2^{-1} \Phi_i^\top \right) \\
		& = \det \left( D_1 - V_1^\top \Phi_i V_2 D_2^{-1} V_2^\top \Phi_i^\top V_1 \right) \\
		& = \prod_{m=1}^b \left( \sigma_{1, \pi_{1, m}} - \frac{\xi_{i, m}^2}{\sigma_{2, \pi_{2, m}}} \right) \, ,
\end{align*}
where $D_1 = \text{diag}(\sigma_{1, 1}, \sigma_{1, 2}, \ldots, \sigma_{1, d_1})$, 
$D_2 = \text{diag}(\sigma_{2, 1},$ $\sigma_{2, 2}, \ldots, \sigma_{2, d_2})$
and $\Xi_i = \text{diag}(\xi_{i, 1}, \xi_{i, 2}, \ldots, \xi_{i, b})$.
Hence, the positive definiteness of $\Lm_i$ is guaranteed if $\xi_{i, m}^2 < \sigma_{1, \pi_{1, m}} \sigma_{2, \pi_{2, m}}$ is satisfied
for $1 \leq m \leq b$.
Moreover, this inequality provides us a guideline on choosing index sets.
Since we want non-zero entries of $\Phi_i$ to be larger, which can be achieved by larger $|\xi_{i, m}|$, 
we choose index sets so that $\sigma_{1, \pi_{1, m}} \sigma_{2, \pi_{2, m}}$ large.
This corresponds to choosing leading eigenvalues and eigenvectors of $\Psi_1$ and $\Psi_2$.
In our simulations, we pick $b = 2$ indices at random from those with eigenvalues in the top $1 / 3$.
We also generate $\xi_{i, m}$ as $\xi_{i, m} = \xi_{0, i, m} \sqrt{\sigma_{1, \pi_{1, m}} \sigma_{2, \pi_{2, m}}}$,
where $\xi_{0, i, m}$ follows a uniform distribution $U([-0.8, -0.5] \cup [0.5, 0.8])$.

For general $a > 2$ cases, we first construct a matrix $\Lm_i^{(1)}$ from $\Psi_1$ and $\Psi_2$.
We then iteratively apply the above procedure to generate $\Lm_i^{(2)}$ from $\Lm_i^{(1)}$ and $\Psi_3$, 
$\Lm_i^{(3)}$ from $\Lm_i^{(2)}$ and $\Psi_4$, until $\Lm_i = \Lm_i^{(a-1)}$ is derived.
In the simulations in Section~\ref{sec:simu}, we set the number of modules to $a = 2$ for $d = 25$, 
$a = 3$ for $d = 50$ and $a = 4$ for $d = 100$.

%==============================================================================================
\section{Proof of Theorems}
\label{app:proof}

%==============================================================================================
\subsection{Proof of Proposition~\ref{prop:cssl_dual}}

Let $E$ and $F_i$ be non-negative $d \times d$ matrices satisfying $- E_{jj'} \leq \Th_{jj'} \leq E_{jj'}$
and $- F_{i, jj'} \leq \Om_{i, jj'} \leq F_{i, jj'}$, respectively, for all $1 \leq i \leq N$ and $1 \leq j, j' \leq d$.
Then, using Lagrange multipliers $\Gamma, \Gamma_0,$ and $\{ \Delta_i, \Delta_{0, i} \}_{i=1}^N$, the CSSL problem (\ref{eq:cssl}) is expressed as
\begin{align*}
	\hspace{-20pt}
	\max_{\Th, E, \{ \Om_i, F_i \}_{i=1}^N } \min_{\Gamma, \Gamma_0, \{ \Delta_i, \Delta_{0, i} \}_{i=1}^N }
		& \sum_{i=1}^N t_i \left\{ \log \det (\Th + \Om_i) - \tr{S_i (\Th + \Om_i)} \right\} \\
	& \hspace{-24pt}- \sum_{j, j' = 1}^d \left\{ \rho E_{jj'} + \gm \left( \sum_{i=1}^N F_{i, jj'}^p \right)^\frac{1}{p} \right\} \\
	& \hspace{-24pt}- \tr{\Gamma \Th} + \tr{{\rm abs}(\Gamma) E} + \tr{\Gamma_0 E} \\
	& \hspace{-24pt} - \sum_{i=1}^N \left\{ \tr{\Delta_i \Om_i} -\tr{{\rm abs}(\Delta_i) F_i} - \tr{\Delta_{0, i} F_i} \right\} \\
	& \hspace{-64pt} \st \Gamma_{0, jj'} \geq 0 \; , \; \Delta_{0, .i, jj'} \geq 0
		\;\; (1 \leq i \leq N \, , \, 1 \leq j, j' \leq d) \, .
\end{align*}
By changing the order of maximization and minimization above, we derive the dual problem.
Now, we optimize each variable $\Th$, $E$, $\Om_i$ and $F_i$ by setting each derivative to zero: 
\begin{align*}
	& \sum_{i=1}^N t_i \left\{ (\Th + \Om_i)^{-1} - S_i \right\} - \Gamma = 0_{d \times d} \; , \\
	& - \rho \Vec{1}_d \Vec{1}_d^\top + {\rm abs}(\Gamma) + \Gamma_0 = 0_{d \times d} \; , \\
	& t_i \left\{ (\Th + \Om_i)^{-1} - S_i \right\} - \Delta_i = 0_{d \times d} 
		\;\; (1 \leq i \leq N) \; , \\
	& - \gm \left( \sum_{i=1}^N F_{i, jj'}^p \right)^\frac{1-p}{p} F_{i, jj'} + \left| \Delta_{i, jj'} \right| + \Delta_{0, i, jj'} = 0 \\
	& \hspace{96pt}(1 \leq i \leq N \, , \, 1 \leq j, j' \leq d) \, .
\end{align*}
As a result of these equations, we get
\begin{align*}
	& \Delta_i = t_i \left\{ (\Th + \Om_i)^{-1} - S_i \right\} \; , \\
	& \left| \sum_{i=1}^N \Delta_{i, jj'} \right| \leq \rho
		\;\; (1 \leq j, j' \leq d) \; , \\
	& \left( \sum_{i=1}^N |\Delta_{i, jj'}|^q \right)^\frac{1}{q} \leq \gm
		\;\; (1 \leq j, j' \leq d) \, .
\end{align*}
and so the dual problem is given by~(\ref{eq:cssl_dual}) where we set $W_i = (\Th + \Om_i)^{-1} = \frac{1}{t_i} \Delta_i + S_i$.
\qed

%==============================================================================================
\subsection{Proof of Theorem~\ref{th:eig_bound}}

We first prove the lower-bound.
Let $W_i = \frac{1}{t_i}\Delta_i + S_i$ in the dual problem~(\ref{eq:cssl_dual}).
Then we have $\left| \sum_{i=1}^N \Delta_{i, jj} \right| \leq \rho$ and $\left( \sum_{i=1}^N |\Delta_{i, jj'}|^q \right)^\frac{1}{q} \leq \gm$, 
and hence 
\begin{align*}
	\snorm{\frac{1}{t_i}\Delta_i + S_i} & \leq \frac{1}{t_i} \snorm{\Delta_i} + \snorm{S_i} \\
	& \leq \frac{d}{t_i} \max_{j, j'} | \Delta_{i, jj'} | + \snorm{S_i} \\
	& \leq \frac{d}{t_i} \max_i \max_{j, j'} | \Delta_{i, jj'} | + \snorm{S_i} \\
	& \leq \frac{d\gm}{t_i} + \snorm{S_i} \, ,
\end{align*}
where the last inequality comes from the general relationship between $\ell_p$-norms 
$\max_i |\Delta_{i, jj'}| \leq \left( \sum_{i=1}^N |\Delta_{i, jj'}|^q \right)^\frac{1}{q}$.
Since $W_i^* = \frac{1}{t_i}\Delta_i^* + S_i = {\Lm_i^*}^{-1}$ holds at the optimum, we have the lower-bound.

We now turn to proving the upper-bound.
From strong duality, the duality-gap is zero at the optimal solution to the primal and the dual problems (\ref{eq:cssl}), (\ref{eq:cssl_dual}), 
and we have
\begin{align*}
	\rho \pnorm{\Th^*}{1} + \gm \pnorm{\Om^*}{1, p} = d - \sum_{i=1}^N t_i \tr{S_i (\Th^* + \Om_i^*)} \, .
\end{align*}
Moreover, from $0 < \rho < N^\frac{1}{p}\gm < \infty$, $\tr{S_i (\Th^* + \Om_i^*)} \geq 0$ and 
the general $\ell_p$-norm rule $\left( \sum_{i=1}^N |\Om_{i, jj'}^*|^p \right)^\frac{1}{p} \geq \max_i |\Om_{i, jj'}^*|$, 
\begin{align*}
	\pnorm{\Th^*}{1} + N^{-\frac{1}{p}} \pnorm{\Om^*}{1, \infty} \leq \frac{d}{\rho}
\end{align*}
holds.
Since $N^\frac{1}{p} \geq 1$ for $p \geq 1$, we get
\begin{align*}
	\pnorm{\Th^*}{1} + \pnorm{\Om^*}{1, \infty} \leq \frac{N^\frac{1}{p} d}{\rho} \, .
\end{align*}
We use this inequality to derive the upper-bound.
From the definition, the precision matrix factorizes as $\Lm_i^* = \Th^* + \Om_i^*$, and hence we have
\begin{align*}
	\snorm{\Lm_i^*} & \leq \snorm{\Th^*} + \snorm{\Om_i^*} \\
	& \leq \snorm{\Th^*} + d \max_{j, j'} | \Om_{i, jj'}^* | \\
	& \leq \snorm{\Th^*} + d \max_i \max_{j, j'} | \Om_{i, jj'}^* | \\
	& \leq \snorm{\Th^*} + d \pnorm{\Om^*}{1, \infty} \\
	& \leq d \left( \snorm{\Th^*} + \pnorm{\Om^*}{1, \infty} \right) \\
	& \leq d \left( \pnorm{\Th^*}{1} + \pnorm{\Om^*}{1, \infty} \right) \\
	& \leq \frac{N^\frac{1}{p} d^2}{\rho}
\end{align*}
Here, we have used the relationship $\snorm{\Th^*} \leq \pnorm{\Th^*}{2} \leq \pnorm{\Th^*}{1}$.
\qed

%==============================================================================================
\subsection{Proof of Theorem~\ref{th:cssl_cond}}

The Hessian matrix of the CSSL primal loss $\sum_{i=1}^N t_i \ell(\Th + \Om_i; S_i)$ is given by
\begin{align*}
	\mathcal{H}_{\rm primal} = - \matrix{ccccc}{
		\sum_{i=1}^N t_i K_i & t_1 K_1 & t_2 K_2 & \ldots & t_N K_N \\
		t_1 K_1 & t_1 K_1 & 0_{d^2 \times d^2} & \ldots & 0_{d^2 \times d^2} \\
		t_2 K_2 & 0_{d^2 \times d^2} & t_2 K_2 & & \vdots \\
		\vdots & \vdots & & \ddots & 0_{d^2 \times d^2} \\
		t_N K_N & 0_{d^2 \times d^2} & \ldots & 0_{d^2 \times d^2} & t_N K_N
	} \, ,
\end{align*}
where $K_i = (\Th + \Om_i)^{-1} \otimes (\Th + \Om_i)^{-1}$.
It is easy to verify that $\Vec{1}_{N + 1} \otimes I_d$ spans a null space of $\mathcal{H}_{\rm primal}$
and thus $\mathcal{H}_{\rm primal}$ is always rank-deficient.

On the other hand, the matrix of the CSSL dual loss $- \sum_{i=1}^N t_i \log \det W_i$ is the block-diagonal matrix
\begin{align*}
	\mathcal{H}_{\rm dual} = \text{block--diag}(t_1 \tilde{K}_1, t_2 \tilde{K}_2, \ldots, t_N \tilde{K}_N) \, ,
\end{align*}
where $\tilde{K}_i = W_i^{-1} \otimes W_i^{-1}$.
From Theorem \ref{th:eig_bound}, we know that the CSSL solution has bounded eigenvalues 
and thus the above Hessian matrix is always strictly positive definite for any feasible $W_i$.
\qed

%==============================================================================================
\subsection{Proof of the Proposition~\ref{prop:hyp_param}}

Let $S_i$ be the covariance matrix $S_i = \matrix{cc}{a_i & r_i \\ r_i & b_i}$.
Then we have an upper-bound for~(\ref{eq:cssl_bivar}) of
\begin{align*}
	& \hspace{-15pt} \sum_{i=1}^N t_i \left\{ \log(u_i v_i - (\theta + \omega_i)^2) - (a_i u_i + b_i v_i + 2 r_i \theta + 2 r_i \omega_i) \right\} \\
	& \hspace{9pt} - 2 \rho |\theta| - 2 \gm \pnorm{\Vec{\omega}}{p} \\
	& \hspace{-15pt} \leq \sum_{i=1}^N t_i \left\{ \log(u_i v_i - (\theta + \omega_i)^2) 
		- (a_i u_i + b_i v_i) - 2 (r_i \omega_i + \gm |\omega_i|) \right\} \\
	& \hspace{9pt} - 2 \left(\sum_{i=1}^N t_i r_i \theta + \rho |\theta| \right) \, ,
\end{align*}
from the relationship $\sum_{i=1}^N t_i |\omega_i| \leq \pnorm{\Vec{\omega}}{\infty} \leq \pnorm{\Vec{\omega}}{p}$.
Moreover, this upper-bound coincides with the original problem when $\Vec{\omega} = \Vec{0}_N$.
Therefore, if $\Vec{\omega} = \Vec{0}_N$ is a maximizer of this upper-bound, it is also a maximizer of~(\ref{eq:cssl_bivar}).
From the derivative of the upper-bound over $\omega_i$, we get that $\omega_i = 0$ is a maximizer if the following condition holds:
\begin{align*}
	- (\gm + r_i) \leq \frac{\theta}{u_i v_i - \theta^2} \leq (\gm - r_i) \, .
\end{align*}
This is a sufficient condition for the original problem~(\ref{eq:cssl_bivar}) to have $\omega_i = 0$ as its solution.
Under this condition, problem~(\ref{eq:cssl_bivar}) can be expressed as
\begin{align*}
	& \max_{\theta, \tilde{u}, \tilde{v}, u_i, v_i} \; \log (\tilde{u} \tilde{v} - \theta^2) 
		- (\tilde{a} \tilde{u} + \tilde{b} \tilde{v}) - 2 (\tilde{r} \theta + \rho |\theta|) \\
	& \hspace{8pt} \st \tilde{u} \tilde{v} - \theta^2 > 0 \; , \\
	& \hspace{24pt} - (\gm + r_i) \leq \frac{\theta}{u_i v_i - \theta^2} \leq (\gm - r_i) \;\; (1 \leq i \leq N)
\end{align*}
for some properly chosen $\tilde{a}, \tilde{b}$ and $\tilde{r} = \sum_{i=1}^N t_i r_i$.
Hence, since the additional condition involves $\theta = 0$ irrelevant to the value of $u_i$ and $v_i$
if $\max_{1 \leq i \leq N} |r_i| \leq \gm$ holds, we have $\theta = 0$ when $|\tilde{r}| \leq \rho$ from \citet[Proposition~1]{ide2009proximity}.
\qed

%==============================================================================================
\subsection{Proof of Theorem~\ref{th:yupdate_c3}}

Let $h(\Vec{y}) = \frac{1}{2} \| \Vec{y} - \Vec{y}_0\|_2^2$ and $\Vec{y}'$ be one of the feasible solutions to the original problem~(\ref{eq:yupdate2}).
Moreover, since $\tilde{\Vec{y}}$ is infeasible for the original problem~(\ref{eq:yupdate2}), $\pnorm{\tilde{\Vec{y}}}{q} > \gm$ holds.
Then, for $\Vec{y}'' = \Vec{y}' + \epsilon (\tilde{\Vec{y}} - \Vec{y}')$ with $0 < \epsilon \leq 1$, 
$h(\Vec{y}'') \leq h(\Vec{y}')$ holds from the convexity of $h$.
Therefore, $\Vec{y}''$ is a better solution to problem~(\ref{eq:yupdate2}) as long as the constraints
$|\Vec{1}_N^\top \Vec{y}''| \leq \rho$ and $\pnorm{\Vec{y}''}{q} \leq \gm$ are satisfied.
The first condition always holds because 
$|\Vec{1}_N^\top \Vec{y}''| \leq (1 - \epsilon) |\Vec{1}_N^\top \Vec{y}'| + \epsilon |\Vec{1}_N^\top \tilde{\Vec{y}}| \leq \rho$.
On the other hand, the latter condition $\pnorm{\Vec{y}''}{q} = \left( \sum_{i=1}^N |y''_i|^q \right)^\frac{1}{q} \leq \gm$
is no longer valid if $\pnorm{\Vec{y}'}{q} = \gm$ and $\sgn{y'_i} = \sgn{\tilde{y}_i - y'_i}$, 
which results in $\tilde{y}_i y'_i \geq 0$.
This is a necessary condition for the solution to~(\ref{eq:yupdate2}).
Otherwise, we can always improve the solution by the above procedure, which contradicts its optimality.
\qed

\bibliographystyle{elsarticle-harv}
\bibliography{nn_cssl}




%% \bibitem must have one of the following forms:
%%   \bibitem[Jones et al.(1990)]{key}...
%%   \bibitem[Jones et al.(1990)Jones, Baker, and Williams]{key}...
%%   \bibitem[Jones et al., 1990]{key}...
%%   \bibitem[\protect\myciteauthoryear{Jones, Baker, and Williams}{Jones
%%       et al.}{1990}]{key}...
%%   \bibitem[\protect\myciteauthoryear{Jones et al.}{1990}]{key}...
%%   \bibitem[\protect\astroncite{Jones et al.}{1990}]{key}...
%%   \bibitem[\protect\mycitename{Jones et al., }1990]{key}...
%%   \harvarditem[Jones et al.]{Jones, Baker, and Williams}{1990}{key}...
%%


% \end{thebibliography}

\end{document}